\PassOptionsToPackage{table}{xcolor}

\documentclass[runningheads]{llncs}

\usepackage{eccv}

\usepackage{eccvabbrv}

\usepackage{graphicx}
\usepackage{booktabs}
\usepackage{microtype}

\usepackage[accsupp]{axessibility}

\usepackage{multirow}
\usepackage{pifont}
\usepackage{array}
\usepackage{siunitx}
\usepackage{makecell}
\usepackage{colortbl}
\usepackage{wrapfig}
\usepackage[most]{tcolorbox}

\usepackage{hyperref}

\usepackage{orcidlink}

\newcommand{\cmark}{\ding{51}}
\newcommand{\xmark}{\ding{55}}
\newcommand{\ourdataset}{\textsc{SVG2}\xspace}
\newcommand{\ourmodel}{\textsc{TraSeR}\xspace}
\newcommand{\ourdatasetfull}{Synthetic Visual Genome 2\xspace}

\definecolor{pastelblue}{RGB}{173,216,230}
\definecolor{pastelpink}{RGB}{255,182,193}
\definecolor{pastelgreen}{RGB}{152,251,152}

\newtcolorbox{pastelbox}[2][]{
    breakable,
    colback=#1!10!white, 
    colframe=#1!80!black,
    boxrule=0.5mm,
    arc=1.5mm,
    auto outer arc,
    left=1mm,
    right=1mm,
    top=1mm,
    bottom=1mm,
    title=#2}

\begin{document}

\title{\ourdatasetfull{}: \\Extracting Large-scale Spatio-Temporal Scene Graphs from Videos}

\titlerunning{Synthetic Visual Genome 2}

\author{Ziqi Gao\inst{1} \and
Jieyu Zhang\inst{1,2} \and
Wisdom Oluchi Ikezogwo\inst{1,2} \and
Jae Sung Park\inst{1} \and
Tario G You\inst{2} \and
Daniel Ogbu\inst{4} \and
Chenhao Zheng\inst{1,2} \and
Weikai Huang\inst{2} \and
Yinuo Yang\inst{2} \and
Winson Han\inst{1} \and
Quan Kong\inst{3} \and
Rajat Saini\inst{3} \and
Ranjay Krishna\inst{1,2}}

\authorrunning{Z.~Gao et al.}

\institute{Allen Institute for AI \and
University of Washington \and
Woven by Toyota \and
Microsoft}
\maketitle

\setcounter{footnote}{0}
\begin{abstract}

We introduce \textbf{\ourdatasetfull{}} (\textbf{\ourdataset{}}), a large-scale panoptic video scene graph dataset.
\ourdataset{} contains over 636K videos with 6.6M objects, 52.0M attributes, and 6.7M relations, providing an order-of-magnitude increase in scale and diversity over prior spatio-temporal scene graph datasets. 
To create \ourdataset{}, we design a fully automated pipeline that combines multi-scale panoptic segmentation, online–offline trajectory tracking with automatic new-object discovery, per-trajectory semantic parsing, and GPT-5-based spatio-temporal relation inference. Human verification of \ourdataset{} annotation accuracy confirms its reliability (objects: 93.8\%, attributes: 88.3\%, relations: 85.4\%).
Building on this resource, we train \ourmodel{}, a video scene graph generation model. \ourmodel{} augments VLMs with a trajectory-aligned token arrangement mechanism and new modules: an object-trajectory resampler and a temporal-window resampler to convert raw videos and panoptic trajectories into compact spatio-temporal scene graphs in a single forward pass. 
The temporal-window resampler binds visual tokens to short trajectory segments to preserve local motion and temporal semantics, while the object-trajectory resampler aggregates entire trajectories to maintain global context for objects.
On the PVSG, VIPSeg, VidOR and SVG2\textsubscript{test} dataset, \ourmodel{} improves relation detection by $+15{\sim}20\%$, object prediction by $+30{\sim}40\%$ over the strongest open-source baselines and by $+13\%$ over GPT-5, and attribute prediction by $+15\%$. When \ourmodel{}'s generated scene graphs are sent to a VLM for \textit{video} question answering, it delivers a $+1.5{\sim}4.6\%$ absolute accuracy gain over using video only or video augmented with Qwen2.5-VL's generated scene graphs, demonstrating the utility of explicit spatio-temporal scene graphs as an intermediate representation\footnote{The SVG2 data, model checkpoints, and code are available at \url{https://uwgzq.github.io/papers/SVG2/}}.
\end{abstract}


\section{Introduction}

\begin{figure}[t]
\centering
\includegraphics[width=0.75\linewidth]{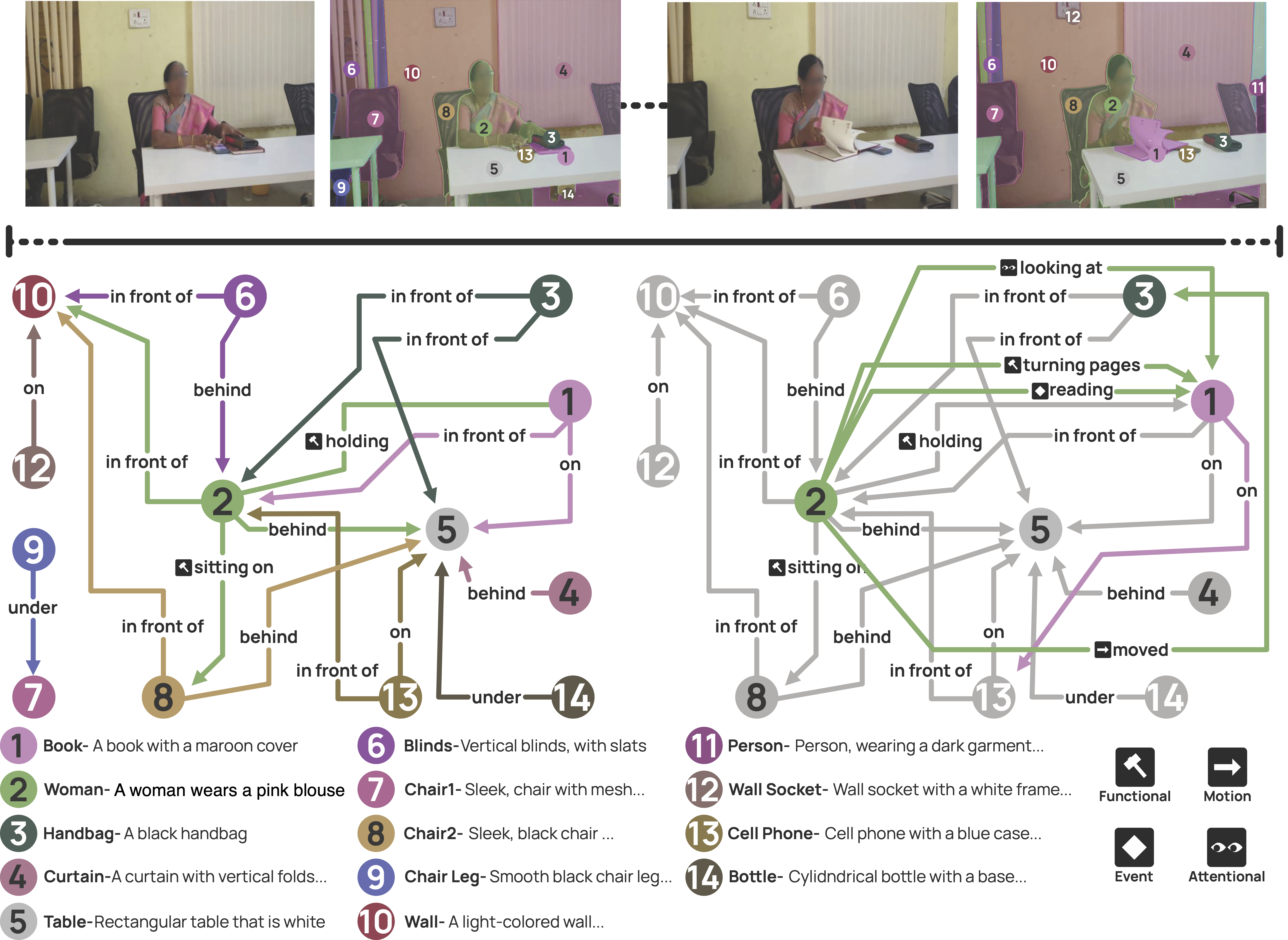}
\caption{\textbf{\ourdatasetfull{} (\ourdataset{})}, a large-scale synthetic panoptic video scene graph dataset.
\ourdataset{} provides dense panoptic trajectories, fine-grained object categories and attributes, and temporally grounded spatialtemporal relations across over 636K videos, which is an order-of-magnitude increase in scale and diversity over prior datasets.}
\label{fig:teaser}
\vspace{-1em}
\end{figure}

Decades of work in human cognition show that people do more than recognizing objects in an image; they rapidly extract structured relational information between those objects~\cite{biederman2017semantics,hummel1992dynamic,ScenePerceptionandUnderstanding,biederman1987recognition}. 
Image scene graphs make this structure explicit and have emerged as a core abstraction in computer vision~\cite{krishna2016visualgenomeconnectinglanguage,johnson2015image}.
A scene graph is a structured, graphical representation of a visual scene, where nodes correspond to objects (e.g., person, bicycle) and their attributes (\eg blue, rectangular, old), and directed edges represent their pairwise relationships (\eg riding). This representation, which bridges the gap between vision and language, has proven to be a powerful intermediate representation~\cite{chang2021scene,farshad2023scenegenie,gao2025generatescenescenegraph,grunde2021agqa,zhang2024provisionprogrammaticallyscalingvisioncentric,huang2025sossyntheticobjectsegments}, improving captioning~\cite{anderson2016spice}, image retrieval~\cite{johnson2015image}, image generation~\cite{johnson2018image}, vision-language evaluation~\cite{ma2023crepe,hsieh2023sugarcrepe,Zhang2024TaskMA}, AI auditing~\cite{kunzel2025visual}, and robotic navigation~\cite{kim2025tacs}.
Spatio-temporal scene graphs~\cite{ji2020actiongenome} expand this abstraction to video reasoning~\cite{gao2022classification,cherian20222}, capturing scene dynamics and parsing events into changes in relationships between objects (\eg riding a bicycle to walking).

Despite their utility, there are very limited high-quality spatio-temporal scene graphs available.
Due to the high cost and difficulty of dense frame-level manual annotation, current datasets remain limited in scale and difficult to expand~\cite{li2025unbiased}.
Attempts to annotate only a handful of frames per video have led to incomplete annotations that do not capture new objects that appear or disappear~\cite{ji2020actiongenome,grunde2021agqa}.
As a result, models trained on such datasets tend to overfit to dataset-specific distributions and predicate classifiers.
This is exacerbated by the severe long-tail bias, sparse and inconsistent annotations, noise, and insufficient temporal labeling~\cite{tang2020unbiased,nag2023unbiasedscenegraphgeneration,qiu2023virtualhome}.
Consequently, scene graph prediction models struggle to generalize to new scenes with an open vocabulary of new objects, attributes, and relationships~\cite{nguyen2025hyperglmhypergraphvideoscene,yang2023panopticvideoscenegraph}.

We address these limitations with \textbf{\ourdatasetfull{} (\ourdataset{}), a large-scale panoptic synthetic video scene graph dataset}. \ourdataset{} contains over \textit{636K} videos, \textit{6.6M} objects, \textit{52M} attributes, and \textit{6.7M} relations 
This dataset is an order-of-magnitude increase in scale and diversity over prior datasets (Table~\ref{tab:dataset_comparison}). 
Human verification of \ourdataset{} confirms the reliability and accuracy of its annotations, yielding 93.8\% accuracy for objects, 88.3\% for attributes, and 85.4\% for relations. To create \ourdataset{}, we design a holistic, scalable, fully automated pipeline for synthesizing high-quality video scene graphs.
Our pipeline unifies multi-scale panoptic segmentation~\cite{sam2}, trajectory tracking with automatic new-object discovery, per-trajectory semantic parsing, and GPT-5-based~\cite{gpt5} spatio-temporal relation inference into a robust video scene graph generation framework.
Our heuristic online-offline two-stage tracking mechanism enables (i) proactively detecting newly emerging instances without human prompts or external detectors and (ii) preserving identity consistency across the entire video.

With \ourdataset{}, \textbf{we design and train \ourmodel{}, a video scene graph parsing model} that converts raw videos and panoptic trajectories into spatio-temporal scene graphs in a single forward pass. Video scene graph parsing is technically challenging for two key reasons. First, object categories are globally stable, but attributes and relations are temporally local and can change rapidly over time. So the model must capture both long-range semantics and fine-grained temporal variation. Second, long videos with many objects produce extremely long token sequences and a combinatorial number of relation candidates, making naive dense encoding and decoding computationally infeasible.

\ourmodel{} addresses these challenges by augmenting VLMs with two modules: a \textit{temporal-window  resampler} and an \textit{object-trajectory resampler}. The temporal-window resampler binds visual tokens to individual object trajectory segments, compacting dense patch-level tokens into short, trajectory-aligned, time-aware streams that preserve relationship and attribute changes. The object-trajectory resampler summarizes across the tokens of a specific object's trajectory to encode global object context.

Coupled with strong pretrained VLM priors and training on \ourdataset{}, \ourmodel{} generates structured, temporally consistent scene graphs. Our resampler designs translate into clear gains: the object-trajectory resampler yields substantial improvements in object recognition, while the temporal-window resampler drives large gains in relation prediction. Empirically, \ourmodel{} improves relation prediction by $+15{\sim}20\%$ over the strongest open-source models~\cite{zeng2024chatglm,hong2025glm,yang2025qwen3}, boosts object prediction by $+30\%$ to $+40\%$ over open-source baselines~\cite{yao2024minicpmv,bai2025qwen25vl,wang2025internvl35,hong2025glm} and by $+13\%$ over GPT-5~\cite{gpt5}, and raises attribute prediction by $+15\%$ relative to open-source state of the art~\cite{hong2025glm,yang2025qwen3,hong2025glm}.

Finally, we evaluate the utility of \ourmodel{}'s generated scene graphs using video question answering.
For this, we use the AGQA dataset~\cite{grunde2021agqa} and the Perception-Test~\cite{patraucean2023perception}.
When we feed \ourmodel{}’s predicted video scene graphs into GPT-4.1, we obtain absolute accuracy gains of $+1.5\%{\sim}4.6\%$ over baselines using video only or video augmented with Qwen2.5-VL's scene graphs.
This result suggests that \ourmodel{} is able to surface and summarize useful information that even Qwen2.5-VL misses.

\begin{figure}[t]
    \centering
    \includegraphics[width=\linewidth]{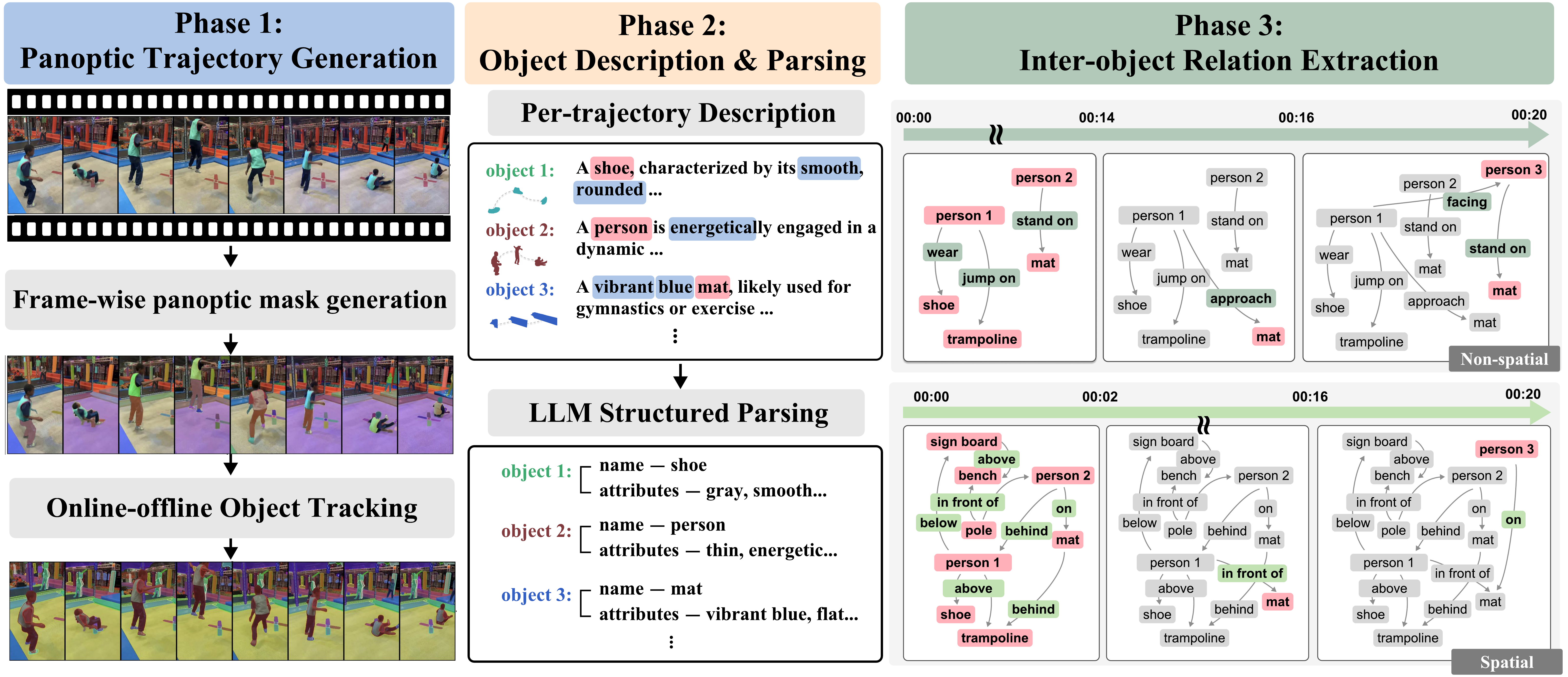}
    \caption{Overview of \ourdataset{} synthesis pipeline. \textbf{Phase 1}: panoptic trajectory generation with online--offline object tracking mechanism that discovers new objects and preserves identity consistency.
    \textbf{Phase 2}: per-trajectory description and semantic parsing.
    \textbf{Phase 3}: GPT-5–based spatiotemporal relation inference to produce the final video scene graph.}
    \label{fig:pipeline}
\end{figure}

\section{Related work}

\noindent\textbf{Scene graph datasets.}
Early scene graph research was enabled by the introduction of Visual Genome~\cite{krishna2016visualgenomeconnectinglanguage}, which provided large-scale object, attribute, and relational annotations. Follow-up datasets focused on improving annotation quality, coverage, or grounding: VRD~\cite{lu2016visualrelationship} offered cleaner predicate labels for relationship detection, Open Images V6~\cite{kuznetsova2020openimages} expanded to mask-grounded relations, and PSG~\cite{yang2022panopticscenegraphgeneration} introduced panoptic segmentation as the grounding for scene graph nodes. Most recently, Synthetic Visual Genome (SVG)~\cite{park2025syntheticvisualgenome} scales image scene graphs to the million-image level and provides dense relational annotations, approximately 4 times more relations per object than Visual Genome.
In the video domain, several Video Scene Graph datasets have been introduced~\cite{shang2017vidvrd,ji2020actiongenome, shang2019annotating,darkhalil2022epickitchensvisorbenchmarkvideo,grauman2022ego4dworld3000hours,rodin2024action,nguyen2024highierarchicalinterlacementgraph}. However, large-scale datasets with dense and temporally structured annotations remain a challenge~\cite{yang2023panopticvideoscenegraph, li2022videoknetsimplestrong}. Our work aims to close these gaps by proposing a robust pipeline to create large-scale video scene graph datasets.

\noindent\textbf{Video scene graph generation.} Benefiting from recent progress in video segmentation and semantic representation~\cite{zheng2026trajtok,zheng2025one,sam2,sam3}, prior efforts in video scene graph generation have attempted to develop unified frameworks capable of constructing temporally grounded object–relation graphs directly from raw video data~\cite{shang2017vidvrd,qian2019video,yang2023panopticvideoscenegraph,nguyen2025hyperglmhypergraphvideoscene,nguyen2024highierarchicalinterlacementgraph,nag2023unbiasedscenegraphgeneration,Teng_2021_ICCV,Cong_2021_ICCV}. Despite these advances, existing approaches exhibit notable limitations. Early methods relying on bounding-box-based scene graphs~\cite{shang2017vidvrd,qian2019video,nguyen2025hyperglmhypergraphvideoscene,nguyen2024highierarchicalinterlacementgraph,nag2023unbiasedscenegraphgeneration} struggle to capture  fine-grained spatial details, particularly for non-rigid objects and amorphous background regions, leading to incomplete or imprecise relational reasoning. More recent work~\cite{yang2023panopticvideoscenegraph,yang20234d} addresses this problem by introducing segmentation-tracking pipelines approach but fails to achieve dense object relationships and attributes.

\section{\ourdatasetfull}
\label{sec:data}
We introduce \ourdataset{}, a large-scale synthetic video dataset built upon our newly designed automated VSG generation pipeline.
Unlike static image scene graphs, video scene graphs must encode temporal dynamics, capturing both the continuous motion of visual entities and the evolving interactions among them. Building such representations requires solving three core challenges: (1) Consistent and accurate instance-level tracking across the entire video, including object emergence and disappearance. (2) Exhaustive identification of all objects and their attributes. (3) Precise extraction and temporal localization of inter-object relationships, covering both spatial and temporal interactions. 
\subsection{Automatic pipeline}
To address these challenges, we propose a fully automated synthesis pipeline (\cref{fig:pipeline}) that integrates SAM2~\cite{sam2}, Describe Anything (DAM)~\cite{lian2025describe}, and GPT-5~\cite{gpt5}, producing dense, temporally grounded, and semantically rich video scene graphs. More details are provided in Appx.~\ref{sec:pipe_details}.

\noindent\textbf{Phase 1: Panoptic trajectory generation.}
To obtain dense and temporally consistent object trajectories, we leverage SAM2 to produce high-quality panoptic masks on each frame using multi-scale grid prompts. These per-frame masks serve as candidates for object tracking. Because SAM2 cannot proactively detect newly appearing objects and is sensitive to prompt initialization, we introduce a two-stage online--offline tracking framework that achieves both dynamic object discovery and global temporal consistency.
In the \emph{online stage}, we propagate initial masks forward while continuously monitoring uncovered regions. When candidate masks cover new or previously untracked areas beyond a threshold, the pipeline identifies object emergence, assigns new object IDs, and resumes propagation. A conservative asymmetric-overlap matching strategy prevents identity switches under occlusion or reappearance.
To restore full temporal history, the \emph{offline stage} replays the video once more, reinitializing each object at its first appearance and tracking all instances jointly in a single forward pass. This produces globally stable, complete trajectories. A lightweight post-filtering module removes redundant overlapping tracks and corrects minor mask artifacts (Appx.~\ref{sec:post-filtering}). 
At IoU = 0.5, our tracking achieves AR of 0.754 (VIPSeg~\cite{miao2022vipseg}), 0.686 (PVSG~\cite{yang2023panopticvideoscenegraph}), and 0.623 (VidOR~\cite{shang2019annotating}), demonstrating robust performance across diverse benchmarks.


\noindent\textbf{Phase 2: Object description and structured parsing.}
After obtaining panoptic segmentation trajectories, we generate detailed textual descriptions for each object track using DAM-3B-Video~\cite{lian2025describe}. These descriptions are then processed by GPT-4.1-nano~\cite{achiam2023gpt} to extract the object name and associated adjectival attributes (e.g., color, shape, etc.). To mitigate synthetic-label noise, we introduce a SAM3-based verification step after Phase 2. We prompt SAM3~\cite{sam3} with object labels extracted from the structured annotations to obtain SAM3 instance trajectories, then match each original trajectory to SAM3 trajectories using video-level spatiotemporal IoU with a Hungarian assignment. SAM3 is used purely as a verifier: we keep the original masks and discard objects whose labels are not supported by the matched SAM3 trajectories.

\noindent\textbf{Phase 3: Inter-object spatiotemporal relation extraction.}
We then pass sampled frames along with object IDs, names, and bounding boxes to GPT-5 for inter-object relation inference. 
Building on prior image scene-graph taxonomies~\cite{ji2020actiongenome,grunde2021agqa}, we define a video-adapted relation space covering seven types: spatial (geometric/positional context), functional (contact and manipulation), stateful (persistent attachment or carrying), motion (relative movement over time), social (animate-to-animate interaction), attentional (gaze or camera focus), and event-level (temporally extended, goal-directed dynamics).
Considering that spatial relations naturally dominate over non-spatial relations in diverse video environments, we split each video into two batches and query GPT-5 separately for spatial and non-spatial relation extraction. For spatial relations, we suppress trivial 2D cues (e.g., left of, right of) derivable directly from mask layout, prompting GPT-5 to focus on non-trivial relations that require visual evidence.

\noindent\textbf{Human verification.}
To assess dataset quality, we randomly sample 
1.2K object trajectories, 2K attributes and 1.1K relations for manual verification. Human annotators report 93.8\% accuracy for objects, 88.3\% for attributes and 85.4\% for relations. Among relation errors, about 86\% stem from predicate errors under occlusion or crowded scenes, and 14\% from temporal range errors.

\subsection{Extracted dataset}
Leveraging our pipeline, we create \ourdataset, which contains 636K automatically annotated videos, and \textbf{SVG2\textsubscript{test}}, a high-quality expert-annotated diagnostic benchmark of 100 videos (More dataset statistics are provided in Appx.~\ref{appdx_stats}).

\begin{table}[t]
\centering
\caption{\textbf{Comparison with related benchmarks.} Subscripts $_S / _D$ denote sparse (sampled) vs. dense (per-frame) annotations. `Cls' indicates the number of categories. (*) AG reports total frame-level relation instances, whereas others report unique trajectory-level instances.}
\label{tab:dataset_comparison}
\scriptsize
\setlength{\tabcolsep}{1.2pt}
\renewcommand{\arraystretch}{0.84}

\begin{tabular}{@{}lccccccccc@{}}
\toprule
\textbf{Data} & \textbf{\#Vid} & \textbf{Ann.} & \textbf{Type} &
\textbf{\begin{tabular}[c]{@{}c@{}}Frame/\\Vid\end{tabular}} &
\textbf{\begin{tabular}[c]{@{}c@{}}\#Obj/\\Traj\end{tabular}} &
\textbf{\begin{tabular}[c]{@{}c@{}}Obj\\Cls\end{tabular}} &
\textbf{\#Rel} &
\textbf{\begin{tabular}[c]{@{}c@{}}Rel\\Cls\end{tabular}} &
\textbf{\#Attr} \\
\midrule
SA-V~\cite{sam2} & 50.9K & \begin{tabular}[c]{@{}c@{}}SAM-2\\+ Human\end{tabular} & Seg\textsubscript{S} & 330 & 0.6M & - & - & - & - \\
VIPSeg~\cite{miao2022vipseg} & 2.8K & Human & Seg\textsubscript{S} & 23 & 38.2K & 124 & - & - & - \\
VidVRD~\cite{shang2017vidvrd} & 0.8K & Human & Box\textsubscript{S} & 304 & 2.4K & 35 & 25.9K & 132 & - \\
AG~\cite{ji2020actiongenome} & 7.8K & Human & Box\textsubscript{S} & 808 & 26.2K & 35 & 1.4M* & 26 & - \\
VidOR~\cite{shang2019annotating} & 7.0K & Human & Box\textsubscript{D} & 1K & 34.6K & 80 & 0.3M & 50 & - \\
PVSG~\cite{yang2023panopticvideoscenegraph} & 338 & Human & Seg\textsubscript{S} & 375 & 6.3K & 125 & 3.6K & 62 & - \\
\midrule
VIPSeg\textsubscript{val}~\cite{miao2022vipseg} & 343 & Human & Seg\textsubscript{S} & 23 & 4.7K & 124 & - & - & - \\
VidVRD\textsubscript{test}~\cite{shang2017vidvrd} & 0.2K & Human & Box\textsubscript{D} & 258 & 0.6K & 35 & 4.8K & 132 & - \\
AG\textsubscript{test}~\cite{ji2020actiongenome} & 1.8K & Human & Box\textsubscript{S} & 679 & 7.9K & 35 & 0.5M* & 26 & - \\
VidOR\textsubscript{val}~\cite{shang2019annotating} & 835 & Human & Box\textsubscript{D} & 1K & 4.0K & 80 & 30.1K & 50 & - \\
PVSG\textsubscript{val}~\cite{yang2023panopticvideoscenegraph} & 62 & Human & Seg\textsubscript{S} & 365 & 1.3K & 108 & 1.0K & 58 & - \\
\midrule
\ourdataset(Ours) & \textbf{636K} & Our pipeline & Seg\textsubscript{D} & 479 & \textbf{6.6M} & \textbf{54.2K} & \textbf{6.7M} & \textbf{35.3K} & \textbf{52M} \\
SVG2\textsubscript{test}(Ours) & 100 & Human & Seg\textsubscript{D} & 160 & 3.2K & 749 & 3.3K & 249 & 9.7K \\
\bottomrule
\end{tabular}
\vspace{-2em}
\end{table}

\noindent\textbf{SVG2.}
We sample \textbf{43K} videos from SA-V~\cite{sam2} and \textbf{593K} videos from PVD~\cite{bolya2025PerceptionEncoder}.  SA-V provides strong manually annotated trajectories, but they are temporally sparse and limited in panoptic completeness. To address this, we introduce a hybrid refinement step during tracking in phase 2: we first propagate the human-annotated masks across all frames, then align and compare the resulting trajectories with those produced by our pipeline. We preserve synthetic masks in regions that lack human annotations. This fusion procedure allows us to retain the precision of manual labels while significantly expanding coverage to achieve dense panoptic trajectories. We then apply our attribute and relation extraction stages, resulting in \textbf{0.66M} object instances, \textbf{5.01M} attributes and \textbf{0.72M} spatiotemporal relations.
For the \textbf{593K} sampled PVD subset, we operate the pipeline fully automatically. We then generate  \textbf{5.89M} object instances, \textbf{46.95M} attributes and \textbf{5.99M} spatiotemporal relations. This large-scale synthetic stage substantially broadens the diversity of \ourdataset (\cref{tab:dataset_comparison}). 

\begin{figure}[t]
\centering
\includegraphics[width=0.9\linewidth]{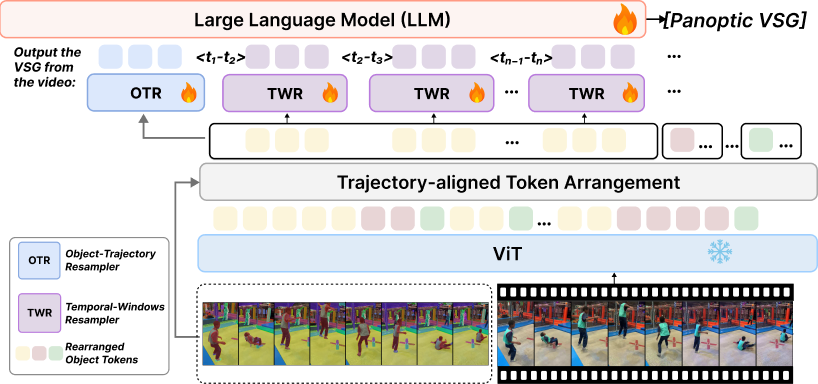}
\caption{\textbf{\ourmodel} architecture. \ourmodel first performs \textbf{trajectory-aligned token arrangement}, grounding ViT tokens to instance trajectories to form identity-preserving token streams. It then applies an \textbf{object-trajectory resampler} to aggregate global semantics over each full trajectory, and a \textbf{temporal-window resampler} to preserve fine-grained motion and temporal cues. The resulting tokens are decoded by the language model into a structured video scene graph.}
\label{fig:model}
\end{figure}

\noindent\textbf{SVG2\textsubscript{test}.}
Existing benchmarks support only closed-set object and relation evaluation~\cite{yang2023panopticvideoscenegraph,shang2019annotating}, with no benchmark jointly assessing objects, attributes and relations over an open vocabulary. We therefore construct SVG2\textsubscript{test}, a high-quality expert-annotated diagnostic benchmark.
We select 100 diverse videos from SA-V~\cite{sam2} and VSPW~\cite{miao2021vspw}.
Reflecting how humans perceive scenes hierarchically, annotators follow a multi-granularity panoptic protocol: coarse instance masks are produced first, then decomposed into salient parts (e.g., a bicycle into frame, wheels, and handlebars), with open-vocabulary categories, multi-class attributes, and time-localized relations assigned and consistency-verified throughout.
As shown in~\cref{tab:dataset_comparison}, this multi-stage, hierarchical labeling process yields substantially denser and more diverse per-video annotations than existing splits (with vocabularies 4$\times$ to 6$\times$ larger and explicit multi-class attributes), enabling evaluation of fine-grained compositional VSG prediction beyond closed-set graphs.



\section{\ourmodel}

The integration of multiple VLMs~\cite{lian2025describe,gpt5,openai2024gpt4omini} in our pipeline ensures high-quality trajectory-level annotations but incurs considerable computational overhead and complexity. Yet, existing VLMs lack native support for panoptic mask trajectory conditioning (typically trained on bbox/point data). To this end, we propose \textbf{\ourmodel}, a trajectory-grounded VLM that produces structured video scene graphs in one pass from raw videos and panoptic object trajectories (\Cref{fig:model}).
Unlike free-form video-language tasks (\eg VideoQA), VSG generation requires accurate alignment between visual semantics and spatio-temporal scene structure, structured prediction of hierarchical graph elements (objects, attributes, relations) rather than natural language, and fine-grained localization with temporal reasoning over video patches to capture interactions, motion, and state changes.
To address these challenges, we first bind video patches directly to object trajectories via trajectory-aligned token arrangement, restoring and preserving consistent object identities over time. We then compress aligned tokens through a dual resampler module: an object-trajectory resampler aggregates global semantic context, and a temporal-window resampler retains fine-grained local motion and temporal dynamics.
\subsection{Trajectory-Aligned Token Arrangement}
We first arrange visual tokens based on object trajectory. 
Given the sampled frame at time $t$, we define $M_{o,t}\in\{0,1\}^{H\times W}$ as the binary segmentation mask for object $o$. 
In ViT, each output token corresponds to a spatiotemporal region obtained by merging $g$ consecutive frames and an $(m \times m)$ grid of base ViT patches (each of size $P \times P$ pixels). Thus, a token indexed by $(t_g, h_m, w_m)$ covers a pixel volume of size $(g\times mP \times mP)$ in the original video.
We compute a coverage score $\mathcal{C}$ for object $o$ at token $(t_g, h_m, w_m)$ by averaging the mask values over the exact pixel support of each token, followed by a temporal max over the frames:
\begin{equation}
\mathcal{C}_{o, t_g, h_m, w_m} = 
\max_{k \in [0, g-1]}
\left(
\operatorname{avgpool}_{mP}(M_{o, t_g \cdot g + k})[h_m, w_m]
\right),
\end{equation}
where $\mathrm{avgpool}(\cdot,mP)$ denotes 2D average pooling on the pixel-level mask. The kernel size and stride are set to $(m \times P)$, matching the token’s exact spatial pixel footprint. Tokens are assigned to object $o$ if their coverage score exceeds a threshold
$\tau_{\mathrm{eff}}\in[0,1]$:
\begin{equation}
I_o = \left\{ (t_g, h_m, w_m) \;\middle|\; \mathcal{C}_{o, t_g, h_m, w_m} \ge \tau_{\mathrm{eff}} \right\}.
\end{equation}
This yields a deterministic set of object-associated visual tokens grounded directly by segmentation evidence. 
After obtaining the set of object-associated tokens, we temporally sort them according to their $t_g$ indices to form an ordered token trajectory for each object. We concatenate these trajectories across all objects, and introduce a special $[\mathrm{TRJ}]$ token to separate objects and explicitly encode trajectory boundaries.
This arrangement produces a structured token stream where each object's trajectory is explicitly segmented and identity-conditioned, providing the LLM with clear object grounding and temporal continuity priors before decoding.
\subsection{Dual Resampler Module}
After trajectory-aligned token arrangement, each object $o$ yields a variable-length token set $X_o \in \mathbb{R}^{|I_o| \times D_{\text{in}}}$.
Directly feeding all tokens into a language model is computationally prohibitive and can bias attention toward densely sampled frames and repeatedly selected tokens, which affects stable training.
To obtain compact yet semantically faithful representations, we introduce two resampler modules that progressively aggregate tokens from pixel-level to object-level and temporal-window-level summaries.

\noindent \textbf{Object-trajectory Resampler.}
Given $X_o$, we introduce $M$ learnable latent queries $L\in \mathbb{R}^{M \times D_{\text{latent}}}$ and resample all arranged tokens of object
$o$ using a Perceiver-Resampler~\cite{alayrac2022flamingo} $f_{\theta}$:
\begin{equation}
    Z_o = f_{\theta}(X_o, L), \qquad Z_o \in \mathbb{R}^{M \times D_{\text{out}}}.
\end{equation}
In this stage, $Z_o$ captures the global semantics of object $o$ over its entire temporal span.

\noindent \textbf{Temporal-window Resampler.}
While object-Level Resampler reduces token count and abstracts global semantics for each video, it may lose time sensitivity, which is crucial for capturing per-object dynamics and inter-object temporal relations.
We therefore partition the video into disjoint windows $\{w\}$ of length $\Delta t$.
Let $X_o^{(w)}$ be the subset of tokens of object $o$ that fall into window $w$.
Using another group of $M'$ latent queries $L'$, we resample each window of each object independently:
\begin{equation}
    Z_o^{(w)} = f_{\theta'}(X_o^{(w)}, L'), \qquad Z_o^{(w)} \in \mathbb{R}^{M' \times D_{\text{out}}}.
\end{equation}
If object $o$ is absent in $w$, no summary is produced for that window, yielding a presence-adaptive, variable-length sequence across time. To expose temporal positions explicitly to the language model, we also add timestamp embeddings before each temporal resampled token.
Finally, the output for object $o$ concatenates the global summary and all window summaries in temporal order:
\begin{equation}
    \mathcal{Z}_o = [Z_o; {Z}_o^{(1)}; {Z}_o^{(2)}; \dots].
\end{equation}

The two resamplers are complementary: object-trajectory Resampler provides an object-centric global semantics, while Temporal-window Resampler restores temporal granularity. The resulting representation $\mathcal{Z}_o$ for each object is compact, object-aware, and time-aware, being well-suited for structured decoding in the language model.
\subsection{Training}\label{sec:training}
\noindent{\textbf{Training datasets.}}
We train \ourmodel by combining our synthetic \ourdataset, which provides complete scene-graph supervision, with established human-annotated benchmarks. Specifically, we integrate video instance segmentation datasets (LV-VIS~\cite{wang2023lvvis}, OVIS~\cite{qi2021ovis}, VIPSeg~\cite{miao2022vipseg}) for robust temporal grounding under occlusion, and video relation datasets (VidOR~\cite{shang2019annotating}, VidVRD~\cite{shang2017vidvrd}) to enrich relational reasoning. To ensure architectural compatibility, we use SAM2 to convert bounding box annotations from VidOR and VidVRD into segmentation masks. Finally, we unify these heterogeneous datasets using task-specific templates that serialize annotations into a standardized autoregressive format, effectively blending dense synthetic structures with complex real-world dynamics.

\noindent{\textbf{Implementation Details.}}
We build on the Qwen2.5-VL-3B and set $\tau_{\mathrm{eff}}{=}0.5$ during trajectory-aligned token arrangement. In the resampling stage, we use two three-layer Perceiver-Resamplers with 32 learnable latent queries each. We finetune the model for one epoch (37K steps), freezing the Qwen2.5-ViT backbone to preserve pretrained visual priors while jointly training the vision–language projector, object-trajectory and temporal-window resamplers, and the language model with learning rates of $5{\times}10^{-5}$, $1{\times}10^{-4}$, and $2{\times}10^{-5}$, respectively. This schedule enables the new resampling modules to adapt more aggressively without destabilizing pretrained components. Models are trained on a single node with 8×H100 GPUs.

\begin{table}[t]
\centering
\caption{We evaluate VLMs on four VidSGG benchmarks using panoptic object trajectories. We report performance across four metrics: triplet recall, relation recall, object accuracy, and attribute recall. The results in this table are reported under the \textbf{lenient} semantic criterion, with a temporal IoU threshold of 0.5 for relations and triplets. \ourmodel outperforms all open-source baselines and surpasses GPT-5 in object and attribute prediction.}
\scriptsize
\setlength{\tabcolsep}{2pt}
\renewcommand{\arraystretch}{0.82}
\resizebox{\textwidth}{!}{
\begin{tabular}{l ccc ccc cccc c c}
\toprule
& \multicolumn{3}{c}{\textbf{Tripet}}
& \multicolumn{3}{c}{\textbf{Relation}}
& \multicolumn{4}{c}{\textbf{Object}}
& \multicolumn{1}{c}{\textbf{Attribute}}
&  \\
\cmidrule(lr){2-4} \cmidrule(lr){5-7} \cmidrule(lr){8-11} \cmidrule(lr){12-12} 
\textbf{Model}
& PVSG & VidOR & SVG2\textsubscript{test}
& PVSG & VidOR & SVG2\textsubscript{test}
& VIPSeg & PVSG & VidOR & SVG2\textsubscript{test}
& SVG2\textsubscript{test}
& \textbf{AvgRank} \\
\midrule
\multicolumn{13}{l}{\textbf{\emph{API call only}}} \\
GPT-4.1~\cite{achiam2023gpt}
& 6.0 & 10.8 & 6.4
& 7.3 & 11.9 & 7.5
& 59.9 & 51.4 & 86.6 & 58.5
& 15.8
& 3 \\
Gemini-2.5 PRO~\cite{comanici2025gemini25}
& {7.4} & 9.8 & 8.7
& {8.8} & 11.0 & 9.9
& 56.7 & 31.7 & 82.9 & 49.8
& 13.6
& 4 \\
GPT-5~\cite{gpt5}
& \textbf{16.6} & \underline{19.7} & \textbf{17.9}
& \textbf{18.3} & \underline{21.7} & \textbf{19.4}
& \underline{68.1} & \underline{54.2} & \underline{88.5} & \underline{65.5}
& \underline{24.1}
& 2 \\
\midrule
\multicolumn{13}{l}{\textbf{\emph{Open weights only}}} \\
Qwen2.5-VL-3B~\cite{bai2025qwen25vl}
& 0.1 & 0.2 & 0.2
& 0.1 & 0.4 & 0.3
& 22.1 & 10.4 & 45.0 & 24.2
& 1.4
& 12 \\
MiniCPM-V 4.5~\cite{yao2024minicpmv}
& 0.1 & 3.0 & 1.1
& 0.2 & 4.0 & 2.4
& 40.0 & 14.3 & 59.1 & 38.5
& 8.4
& 8 \\
InternVL3.5-4B~\cite{wang2025internvl35}
& 0.2 & 0.4 & 0.1
& 0.3 & 0.5 & 0.2
& 33.7 & 20.4 & 66.4 & 35.0
& 7.4
& 11 \\
GLM-4.1-9B-Thinking~\cite{zeng2024chatglm}
& 0.3 & 3.9 & 1.8
& 0.5 & 5.0 & 2.9
& 46.5 & 17.8 & 61.1 & 28.5
& 9.1
& 6 \\
Qwen3-VL-4B~\cite{yang2025qwen3}
& 0.1 & 0.7 & 1.4
& 0.1 & 0.8 & 1.6
& 34.1 & 21.8 & 65.8 & 35.6
& 8.3
& 10 \\
Qwen3-VL-4B-Thinking~\cite{yang2025qwen3}
& 0.1 & 2.3 & 3.3
& 0.4 & 3.4 & 3.6
& 35.8 & 18.3 & 67.6 & 37.1
& 8.8
& 7 \\
\midrule
\multicolumn{13}{l}{\textbf{\emph{Fine-tuned baselines}}} \\
FT-Qwen2.5-VL-3B (First Bbox)
& 0.1 & 1.6 & 0.1
& 0.9 & 4.5 & 0.5
& 25.5 & 25.9 & 51.7 & 36.7
& 10.4
& 9 \\
FT-Qwen2.5-VL-3B (Bbox Traj.)
& 0.5 & 1.8 & 1.4
& 1.6 & 4.2 & 3.0
& 35.1 & 33.6 & 56.9 & 46.1
& 13.4
& 5 \\
\midrule
\ourmodel &
\underline{16.1} & \textbf{22.9} & \underline{16.7} &
\underline{16.9} & \textbf{25.0} & \underline{18.7} &
\textbf{86.5} & \textbf{72.7} & \textbf{91.4} & \textbf{79.0} &
\textbf{27.1} &
1 \\
\bottomrule
\end{tabular}
}
\vspace{-2.5em}
\label{tab:baseline}
\end{table}

\section{Experiment}\label{sec:experiment}
We compare our model against various proprietary and open-source VLM baselines using a standardized setup in which all models receive identical object trajectories and structured prompting templates.
We evaluate open-vocabulary video scene graph generation, assessing object, attribute, relation, and triplet prediction across multiple academic benchmarks and our $SVG2_{test}$. We introduce an LLM-based judge that enables open-vocabulary evaluation through semantic, hierarchy-aware matching, overcoming the closed-set constraints of traditional metrics.
Finally, we perform extensive ablations on training data composition and architectural components to isolate their individual contributions.

\noindent{\textbf{Baselines.}}
We compare \ourmodel with leading proprietary and open-source VLMs, including Gemini-2.5-Pro~\cite{comanici2025gemini25}, GPT-4.1~\cite{achiam2023gpt}, GPT-5~\cite{gpt5}, Qwen-3-VL-4B~\cite{wang2024qwen2vl}, InternVL-3.5-4B~\cite{wang2025internvl35}, GLM-4.1-9B-Thinking~\cite{zeng2024chatglm}, and MiniCPM-4.5~\cite{yao2024minicpmv}.
Since existing VLMs cannot effectively process panoptic mask trajectories, we supply the baselines with dense bounding box trajectories to maximize their supported spatial-temporal context.
We construct instruction-aligned prompting templates for baselines and require all models to output structured scene-graph predictions following a JSON schema. 
To disentangle the benefits of our architecture from the data effects, we fine-tune Qwen2.5-VL-3B on either first-appearance or full bounding-box trajectories, determining whether standard fine-tuning alone is sufficient to generate high-quality VSGs.

\noindent{\textbf{Evaluation Setup.}}
We evaluate video scene graph generation under the open-vocabulary (OV-VSGGen) setting, where the model generates a complete scene graph from ground-truth object trajectories. This entails four prediction sub-tasks: (1) object (labeling each trajectory), (2) attribute (assigning associated characteristics to objects), (3) relation (detecting interactions between trajectories), and (4) triplet (jointly identifying subject-relation-object pairs).

We evaluate on PVSG~\cite{yang2023panopticvideoscenegraph}, VidOR~\cite{shang2019annotating}, VIPSeg~\cite{miao2022vipseg}, and our $\mathrm{SVG2}_{\mathrm{test}}$. While VIPSeg is limited to object prediction and PVSG/VidOR focus on objects and relations, Our $\mathrm{SVG2}_{\mathrm{test}}$ provides complete object–attribute–relation annotations, enabling holistic VidSGG evaluation.

\begin{wraptable}{r}{0.7\textwidth}
\centering
\scriptsize
\vspace{-3em} 
\caption{Validation of the LLM semantic aligner. We report Cohen's $\kappa$ agreement scores between GPT-4o-mini, Gemini 3 Flash  and human annotators. The high agreement confirms the reliability of the automated aligner. ``Id+Syn'' denotes \textit{Identical} or \textit{Synonym} matches; ``Lenient'' includes all valid semantic categories.}
\label{tab:eval_metric}
\resizebox{\linewidth}{!}{%
\begin{tabular}{@{}lcccc@{}} 
\toprule
 & \multicolumn{2}{c}{\textbf{Object}} & \multicolumn{2}{c}{\textbf{Relation}} \\
\cmidrule(lr){2-3} \cmidrule(l){4-5}
& \textbf{Id+Syn} & \textbf{Lenient} & \textbf{Id+Syn} & \textbf{Lenient} \\
\midrule
\makecell[l]{\textbf{Human} vs.\\ \textbf{GPT-4o-mini}} & 0.901 & 0.877 & 0.842 & 0.791 \\
\addlinespace
\makecell[l]{\textbf{Gemini 3 Flash} vs.\\ \textbf{GPT-4o-mini}} & 0.846 & 0.724 & 0.750 & 0.765 \\
\bottomrule
\end{tabular}%
}
\vspace{-2.5em} 
\end{wraptable}

Since VLMs perform open-vocabulary autoregressive generation, traditional closed-set metrics such as Recall@\textit{k} become inadequate. Standard metrics may unfairly penalize valid predictions due to lexical diversity (e.g., human vs. person). Because curating a manual synonym dictionary is impossible for unbounded vocabularies, we introduce a two-tiered evaluation framework.

First, we compute a rule-based \textbf{strict score} requiring exact string matches between ground-truth and predicted elements. Second, to accurately reflect the model's true semantic understanding without unfairly penalizing language diversity, we compute a \textbf{lenient score} using an LLM-assisted semantic aligner. To prevent the self-referential bias of free-form grading, we strictly constrain GPT-4o-mini~\cite{openai2024gpt4omini} to act purely as a pairwise lexical matcher. 
Guided by a rigidly structured prompt, it classifies ground-truth and predicted elements into five categories: \textit{Identical}, \textit{Synonym}, \textit{Hypernym/Hyponym}, \textit{Semantic Overlap}, or \textit{Mismatch}. Any category other than \textit{Mismatch} is considered a correct prediction in lenient settings.

This dual-metric framework applies to objects, attributes, and relations.
Relations additionally require strict temporal grounding (interval IoU > threshold). Triplets are only considered correct if the subject, object, and relation all meet the semantic match criteria and the temporal intersection rules. This evaluation approach ensures a rigorous, bias-mitigated assessment of open-vocabulary scene graphs.

To validate our automated aligner (GPT-4o-mini), we assess its alignment with human evaluators and an independent model (Gemini 3 Flash). Evaluating a stratified subset of 250 objects and 150 relations, we observe substantial inter-annotator agreement (Cohen’s $\kappa$) between the judge and humans (\cref{tab:eval_metric}). This confirms that our constrained matching approach effectively mitigates hallucination, establishing the automated metric as a highly reliable proxy for human judgment.

\begin{table}[t]
\centering
\caption{Training-data ablation study. Incorporating SVG2 improves triplet and object prediction, and boosts object accuracy and attribute recall over academic-only training. Training with the full configuration achieves the best overall performance.}
\label{tab:training_data}
\scriptsize
\setlength{\tabcolsep}{3pt}
\renewcommand{\arraystretch}{1.05}
\resizebox{\textwidth}{!}{
\begin{tabular}{l ccc ccc cccc c c}
\toprule
\textbf{Training}
& \multicolumn{3}{c}{\textbf{Triplet}}
& \multicolumn{3}{c}{\textbf{Relation}}
& \multicolumn{4}{c}{\textbf{Object}}
& \multicolumn{1}{c}{\textbf{Attr.}}
& \textbf{Rank} \\
\cmidrule(lr){2-4}\cmidrule(lr){5-7}\cmidrule(lr){8-11}\cmidrule(lr){12-12}
& PVSG & VidOR & SVG2\textsubscript{t}
& PVSG & VidOR & SVG2\textsubscript{t}
& VIPSeg & PVSG & VidOR & SVG2\textsubscript{t}
& SVG2\textsubscript{t}
& \\
\midrule
Acad                 &  1.35 & 13.61 &  2.68 & 12.55 & 28.66 & 18.74 & 32.71 & 21.50 & 61.09 & 25.68 &  --   & 4 \\
SA-V                 &  8.29 & 15.84 & 13.58 &  9.86 & 17.12 & 16.39 & 81.61 & 72.92 & 90.52 & 77.64 & 25.96 & 3 \\
Acad+SA-V            & 10.70 & 19.23 & 13.00 & 11.12 & 21.06 & 14.69 & 83.10 & 73.46 & 90.96 & 78.14 & 22.72 & 2 \\
Acad+SA-V+PVD        & 16.91 & 22.87 & 16.63 & 16.13 & 25.01 & 18.74 & 86.53 & 72.74 & 91.42 & 79.01 & 27.10 & 1 \\
\bottomrule
\end{tabular}
}
\end{table}



\subsection{Scene graph generation results.}
Consistent with our evaluation framework, \Cref{tab:baseline} reports model performance under the \textbf{lenient} semantic criterion with a temporal IoU threshold of 0.5. Results under the \textbf{strict} exact-match score and a relaxed IoU threshold are detailed in Appx.~\ref{appx:more_results}.
Our light-weight model achieves strong improvements across all benchmarks (\cref{tab:baseline}), outperforming open-source baselines by substantial margins, approximately \textbf{+15\%} in triplet detection, \textbf{+15\%} in relation detection, \textbf{+35\%} in object prediction, and \textbf{+15\%} in attribute recognition. \ourmodel also surpasses proprietary models such as GPT-4.1 and Gemini-2.5-Pro, and even outperforms GPT-5 on object and attribute prediction, highlighting the effectiveness of our model design and training data. We further observe a pronounced performance gap between open-source and proprietary VLMs, especially on longer and more complex videos such as those in PVSG that require precise temporal localization of dynamic relations. This underscores the importance of our large-scale, high-quality open video scene graph dataset (\ourdataset) that provides the rich temporal and dense object–attribute–relation annotations. Both fine-tuned baselines still underperform \ourmodel, confirming that our performance gains stem from our object-centric temporal design rather than mere task-specific fine-tuning. Standard fine-tuning alone is insufficient to capture temporal scene dynamics. As detailed in Appx.~\ref{appx:more_results}, \ourmodel maintains its substantial lead over all baselines even when evaluated under the rule-based strict score and a lower IoU threshold.

\subsection{Video question answering with scene graphs.}
We further assess the role of structured video scene graphs in video QA. We sample 1K open-ended questions from the AGQA 2.0 benchmark~\cite{grunde2021agqa} and 500 multiple-choice questions from the Perception-Test benchmark~\cite{patraucean2023perception}. For each associated video, we first use our pipeline (Phase 1) to generate panoptic object trajectories, and then predict the corresponding video scene graphs using both our trained \ourmodel and the baseline Qwen2.5-VL-3B model.

\begin{wraptable}{r}{0.55\textwidth}
\vspace{-2em} 
\centering
\scriptsize
\caption{GPT-4.1 VQA performance on AGQA 2.0 and Perception-Test. Using \ourmodel's video scene graphs consistently improves performance over baselines.}
\label{tab:gpt4-vqa}
\setlength{\tabcolsep}{3pt}
\renewcommand{\arraystretch}{1.05}
\resizebox{\linewidth}{!}{%
\begin{tabular}{@{}lcc@{}}
\toprule
\textbf{Input} & \textbf{AGQA 2.0} & \textbf{Perception-Test} \\
\midrule
video-only & 25.9 & 66.8 \\
\addlinespace
\makecell[l]{Video+\\QWen2.5-VL-3B's VSG} & 24.8 & 68.6 \\
\addlinespace
\makecell[l]{Video+\\\ourmodel's VSG} & \textbf{26.3} & \textbf{71.4} \\
\bottomrule
\end{tabular}%
}
\vspace{-2em} 
\end{wraptable}
During testing, we sample video frames at 1 fps and feed them into GPT-4.1 under three conditions: (1) video-only input, (2) video plus scene graphs generated by Qwen2.5-VL, and (3) video plus scene graphs generated by our \ourmodel. The results, summarized in~\cref{tab:gpt4-vqa}, show that incorporating high-quality video scene graphs (\eg those generated by \ourmodel) reliably improves VideoQA accuracy (+0.4\% on AGQA 2.0 and +4.6\% on Perception Test). We also conduct a blind-video evaluation on AGQA. We instruct a blind LLM (GPT-4.1-mini) to answer questions using \textit{only} the text-based scene graph, without visual input. \ourmodel's graphs achieve 13.22\% accuracy, outperforming Qwen2.5-VL's graphs by +6.5\%.
We attribute this gain to the structured and fine-grained spatiotemporal information captured by the video scene graphs, which complements the raw visual input and allows the VLM to reason more effectively.


\subsection{Ablation Study}


\noindent\textbf{Training data Ablations.}
We compare four training configurations: academic datasets only (Acad) (\cref{sec:training}), the SVG2 SA-V subset (SA-V), their combination (Acad + SA-V), and the full setup adding the PVD subset (Acad + SA-V + PVD). 
Since academic datasets lack attribute annotations, we evaluate models trained on them only for object and relation outputs.
As shown in~\cref{tab:training_data}, the full configuration achieves the best overall performance. Training solely on Acad overfits to specific benchmarks like VidOR. This results in poor generalization to PVSG and SVG2\textsubscript{test}, alongside low object accuracy.
Incorporating SA-V and PVD yields substantial gains, demonstrating that our dataset's dense, temporally aligned annotations, diverse categories, and visual variability are crucial for learning structured scene graphs.
\begin{table}[t]
\centering
\scriptsize
\caption{Model architecture ablation study. Combining the object-trajectory and temporal-window resamplers yields the best VSG generation performance. Segmentation-aligned token arrangement further provides superior grounding compared to bbox-based signal. \textbf{(*)}Variants without resamplers cannot support long-video inference; therefore, PVSG results are reported on the PVSG VidOR subset only.}
\label{tab:ablation}
\resizebox{0.9\textwidth}{!}{
\begin{tabular}{ccccccccc}
\hline
\multicolumn{3}{c}{Variants} &
\multicolumn{2}{c}{\textbf{Triplet}} &
\multicolumn{2}{c}{\textbf{Relation}} &
\multicolumn{2}{c}{\textbf{Object}} \\
\cmidrule(lr){1-3} \cmidrule(lr){4-5} \cmidrule(lr){6-7} \cmidrule(lr){8-9}
\begin{tabular}{@{}c@{}}Temporal-window\\resampler\end{tabular} &
\begin{tabular}{@{}c@{}}Object-trajectory\\resampler\end{tabular} &
Supervision &
PVSG* & VidOR & PVSG* & VidOR & PVSG* & VidOR \\
\hline
\xmark & \xmark & Seg. & 7.03 & 13.56 & 7.15 & 14.58 & 73.34 & 90.79 \\
\xmark & \cmark & Seg. & 2.44 & 8.23  & 2.68 & 9.04  & 77.10 & 89.62 \\
\cmark & \xmark & Seg. & 8.52 & 15.00 & 8.63 & 16.78 & 77.84 & 89.76 \\
\hline
\cmark & \cmark & Bbox & 8.69 & 15.01 & 9.39 & 15.19 & 73.05 & 89.76 \\
\cmark & \cmark & Seg. & 10.14 & 15.02 & 11.38 & 16.30 & 78.21 & 90.15 \\
\hline
\end{tabular}
}
\end{table}

\noindent\textbf{Model ablations.}
\Cref{tab:ablation} reports ablations on our architectural choices, trained for three epochs on the SVG2 SA-V subset. Using only the object-trajectory resampler eases model training but collapses temporal grounding, while the temporal-window resampler is essential for accurate dynamic relation localization. Combining both yields the best balance of object-level and temporal alignment. We also compare supervision signals for token arrangement: replacing segmentation trajectories with bounding boxes significantly degrades performance on PVSG, as longer, more dynamic videos make coarse box signals too noisy for stable grounding.


\section{Limitations and Discussion}
While \ourdataset{} inherits the current boundaries of the automated models used to synthesize it, this approach unlocks a scale, density, and consistency that is practically unattainable through manual annotation. Importantly, it initiates a virtuous cycle: as foundation models naturally advance, they can be continuously leveraged to refine and elevate future iterations of the dataset.
Architecturally, \ourmodel is currently designed as a trajectory-grounded model. A compelling direction for future work is to extend this approach into a unified, fully end-to-end framework capable of natively proposing and tracking objects alongside VSG generation. Finally, spatio-temporal scene graphs remain a foundational representation for complex video reasoning. With the introduction of \ourdataset{}, we see a promising opportunity to integrate large-scale, structured VSG data directly into the training pipelines of general VLMs, potentially unlocking much deeper compositional and temporal understanding.

\section{Conclusion}
We present \textbf{\ourdatasetfull{}} (\textbf{\ourdataset{}}), a large-scale synthetic panoptic video scene graph dataset of over 636K videos with dense object, attribute, and relation annotations. \ourdataset{} is an order-of-magnitude expansion over prior resources, which was built entirely through an automated pipeline. Leveraging this dataset, we introduce \ourmodel{}, which augments a VLM with object-trajectory and temporal-window resamplers to produce complete video scene graphs in a single forward pass.


\section*{Acknowledgements}
This work is funded by Toyota Motor, Inc.

\bibliographystyle{splncs04}
\bibliography{main}

\newpage
\appendix
\section{Pipeline Details}
\label{sec:pipe_details}

\subsection{Panoptic Trajectory Generation} 
\subsubsection{Frame-wise panoptic mask generation}

We first employ SAM2~\cite{sam2} to automatically produce initial per-frame panoptic segmentation masks. SAM2 is selected for its strong generalization across diverse visual domains and its ability to generate high-quality masks without task-specific tuning. To ensure comprehensive spatial coverage, we apply a multi-grid point prompting strategy using three grids (32×32, 16×16, and 4×4), enabling the model to capture both large regions and fine details. Table~\ref{tab:sam2-automatic-mask-generator-params} shows key parameters that are configured to balance segmentation quality and runtime efficiency.
\begin{table}[] 
\centering
\caption{Configuration of the \textit{SAM2AutomaticMaskGenerator} employed in Phase 1 to produce dense per-frame segmentation proposals.}
\label{tab:sam2-automatic-mask-generator-params}
\scriptsize 
\setlength{\tabcolsep}{6pt} 
\renewcommand{\arraystretch}{1.1} 
\begin{tabular}{ll}
\toprule
\textbf{Parameter} & \textbf{Value} \\
\midrule
\textbf{point\_grids} & $\{32{\times}32,\ 16{\times}16,\ 4{\times}4\}$ \\
\textbf{points\_per\_batch} & 64 \\

\textbf{pred\_iou\_thresh} & 0.75 \\
\textbf{stability\_score\_thresh} & 0.85 \\
\textbf{stability\_score\_offset} & 1.0 \\

\textbf{box\_nms\_thresh} & 0.7 \\
\textbf{crop\_n\_layers} & 2 \\
\textbf{crop\_nms\_thresh} & 0.7 \\
\textbf{crop\_overlap\_ratio} & 0.5 \\

\textbf{min\_mask\_region\_area} & 200 \\
\textbf{use\_m2m} & True \\
\textbf{multimask\_output} & False \\
\bottomrule
\end{tabular}
\end{table}

However, the raw SAM2 outputs often contain redundant or highly overlapping masks, including fragmented sub-object regions and densely stacked proposals that inflate computational cost and complicate downstream tracking. To mitigate these issues, we introduce an optional coverage-optimized mask filter. All candidate masks are first sorted by area, favoring large, semantically coherent regions. Each mask is retained only if its overlap with the already selected set is below a 90\% threshold, which empirically balances redundancy removal and coverage preservation. Filtering continues until the union of selected masks reaches the pre-computed full coverage of all SAM2 proposals.

This filtering step significantly reduces mask quantity while preserving essential scene content, thereby lowering memory usage and improving the stability and efficiency of subsequent tracking stages.

\subsubsection{Online--offline Object Tracking:}
After generating candidate masks on frames, we apply a online--offline object tracking mechanism. This design tackles key challenges in video object tracking: it enables the detection of newly appearing objects during tracking, ensures temporal consistency of segmentation masks, and maintains dense annotations across frames. By dynamically adapting to scene changes, this mechanism enhances the continuity and accuracy of object-level representations throughout the video.

\noindent{\textbf{Online Tracking.}}
The online tracking stage initializes object tracking by loading the filtered candidate masks ($\mathcal{M}'$) from the first frame and registering them with the SAM2 video predictor, denoted by the online tracking function $\mathcal{F}_\text{online}$:
\begin{equation}
    \mathcal{T}_{1}, \mathcal{R} = \mathcal{F}_\text{online}(\mathcal{M}') \,.
    \label{eq:1}
\end{equation}
Here, $\mathcal{T}_{1}$ denotes tracked masks from the first pass, and $\mathcal{R} = {(id_j, t_j, m_j)}$ records the initial appearance of each object, including its identifier $id_j$, the frame of first appearance $t_j$, and the initial mask $m_j$.
Each mask is assigned a unique object identifier that will be maintained throughout the tracking process.

The SAM2 video predictor then propagates these masks across subsequent frames. Although SAM2 provides state-of-the-art video segmentation through memory attention and large-scale pretraining, it heavily depends on manually provided visual prompts such as points, bounding boxes, or masks. As a result, SAM2 lacks the capability to autonomously discover and incorporate newly appearing objects into its memory during propagation. Extensions such as Grounded SAM2~\cite{ren2024grounded} have explored integrating open-vocabulary object detectors to identify new objects in videos. However, these methods often disrupt tracking consistency by failing to maintain consistent object identities and operate only locally, which can compromise existing tracking information.

To enable reliable discovery of new objects, we introduce a key innovation in the online object tracking: a dynamic mechanism for continuous monitoring of tracking quality and the emergence of new objects. 
For each frame t, the set of currently tracked objects is
\begin{equation}
    \mathcal{O}_t = \{ (id_j, m_{t,j}) \mid \text{object } id_j \text{ is tracked at frame } t \} \,,
    \label{eq:2}
\end{equation}
where $m_{t,j}$ denotes the segmentation mask of object $id_j$. The pipeline then computes their combined coverage
\begin{equation}
C_t = \bigvee_{(id_j, m_{t,j}) \in \mathcal{O}t} m_{t,j} \,,
\end{equation}
and the untracked region is $U_t = \neg C_t$, where $\neg$ denotes the logical NOT operation applied pixel-wise. Meanwhile, the union of candidate masks from the automatic generation stage is
\begin{equation}
D_t = \bigvee_{m \in \mathcal{M}'_t} m \,,
\end{equation}
and the overlap between untracked and newly detected areas is
\begin{equation}
O_{\text{overlap}} = U_t \land D_t \,.
\end{equation}


To determine whether a significant portion of the untracked region is explained by these candidate masks, we compute the normalized overlap ratio:
\begin{equation}
\frac{A(O_{\text{overlap}})}{A(U_t)} \geq \tau_{\text{detection}}, \quad \text{where } \tau_{\text{detection}} = 0.1 \,,
\end{equation}
where $A(\cdot)$ denotes the area of a mask. If this ratio exceeds a predefined detection threshold $\tau_{detection}$, the pipeline interprets this as a tracking breakpoint, suggesting that new objects have entered the scene or that existing objects have undergone substantial changes in appearance.

At this breakpoint, forward propagation temporarily halts, and the pipeline evaluates the candidate masks in $\mathcal{M}'_t$. Masks that exhibit high overlap with $U_t$ and do not correspond to any currently tracked object are assigned new unique identifiers and appended to the tracking state $\mathcal{R}$. This enables us to dynamically expand the tracking scope, ensuring the integration of objects that emerge mid-sequence. Importantly, the appearance of each new object is recorded precisely, capturing the frame index $t$ along with its initial segmentation mask. This object entry record supports the offline tracking stage, which further refines tracking across the entire video.

After incorporating new objects, the pipeline performs identity matching to maintain temporal consistency. For each newly added mask $m_{\text{new}}$, we compute its overlap with all existing tracked masks, using an asymmetric overlap ratio defined as:
\begin{equation} 
r_j = \frac{\sum_{x,y} \left( m_{\text{new}} \land m_{t,j} \right)(x, y)}{A(m_{\text{new}})} \quad \forall (id_j, m_{t,j}) \in \mathcal{O}_t \,.
\end{equation}
Unlike IoU, this overlap metric is asymmetric and specifically measures how much of the new object is covered by an existing one, which is more robust in handling partial occlusion scenarios where smaller objects may only be partially visible.

Next, the pipeline identifies the tracked mask with the maximum overlap
\begin{equation}
id_{\text{assigned}} =
\begin{cases}
id_{j_{\max}} & \text{if } r_{\max} \ge \tau_{\text{match}}, \\
\displaystyle
\max_{(id_j, m_{t,j}) \in \mathcal{O}_t} id_j + 1
& \text{otherwise,}
\end{cases}
\end{equation}
where $r_{max} = \max r_j$. If sufficient overlap exists, the mask continues an existing trajectory; otherwise, a new object ID is assigned. This conservative matching strategy prevents identity switches and maintains consistent IDs even under occlusion, reappearance, or new-object entry.

\noindent{\textbf{Offline Tracking.}}
During the online tracking phase, new objects are only discovered and registered when specific breakpoint conditions are met. Because these breakpoints are evaluated periodically, a newly appearing object may not be registered at the exact moment it enters the scene, resulting in delayed initialization and incomplete trajectories prior to its registration. To address this, the offline propagation stage leverages the complete object entry information collected during the online pass to generate refined, full-length tracks:
\begin{equation} 
\mathcal{T} = \mathcal{F}_\text{offline}(\mathcal{R})\,.
\end{equation}

This propagation begins by resetting the SAM2 tracking state. Each object recorded in the online pass is re-initialized at its true entry frame using its initial mask.
This enables all objects to be tracked from their correct starting points in a single continuous pass, ensuring consistent identities across the full sequence. Unlike the online propagation, which may halt at detection breakpoints, the offline propagation performs uninterrupted propagation through all frames for maximal stability and efficiency.
On 100 sampled SVG2 videos, online-only propagation attains mask coverage 0.435, while online+offline improves to \underline{0.486}. 

\noindent{\textbf{Lightweight Post-filtering.}}\label{sec:post-filtering}
We apply a lightweight post-filtering step to refine the panoptic trajectories produced in Phase 1. Given the set of instance mask trajectories, we first remove redundant tracks by detecting near-duplicates through trajectory-level overlap consistency: pairs of tracks that exhibit sustained, high overlap over their co-visible frames are treated as duplicates, and only the more stable trajectory is retained. We then apply a mild morphological cleanup to the remaining masks to suppress small artifacts and smooth boundaries. This post-filtering yields a more compact and cleaner set of trajectories for downstream VSG construction.


\subsection{Object Description and Parsing}

We adopt the Describe Anything Model (DAM)~\cite{lian2025describe} to generate detailed semantic descriptions for each object trajectory. DAM is specifically designed for localized vision–language understanding: given a region of interest, it produces fine-grained descriptions grounded strictly on that region mask while still leveraging global scene context. Compared to generic VLMs that focus on whole-image captioning, DAM is trained with a large-scale Detailed Local Captioning dataset, enabling it to reliably describe small, partially occluded, or visually complex objects. This makes DAM particularly well aligned with our goals, where each tracked trajectory requires accurate semantics.

We use the DAM-3B-Video variant. For each object trajectory, we compute the mask area over its full lifetime, select the top 8 frames with the largest visible regions, and order them temporally. These frames and their binary masks form a short masked clip that is fed into DAM-3B-Video, along with an instruction prompt requesting a detailed description. The resulting output provides high-quality, fine-grained textual descriptions that serve as input to our structured parsing module.

To convert DAM’s free-form descriptions into structured semantics, we use GPT-4.1-nano~\cite{achiam2023gpt} for object name and attribute list parsing, balancing quality and cost. The parsing prompt is shown in Table~\ref{tab:prompt_parse}.
In practice, we find that asking the model to also output relations, motions, and other elements helps it separate concepts more clearly, resulting in a cleaner and more accurate attribute list.

\noindent{\textbf{API Usage.}} We use \texttt{gpt-4.1-nano-2025-04-14} for object name and attribute list parsing, processing $\sim10B$ input tokens and generating $\sim60B$ output tokens.

\subsection{Inter-object Relation Extraction}
After Phase 1 and Phase 2, we obtain panoptic object trajectories and an object name for each trajectory. To further exploit the reasoning and grounding capabilities of VLMs, we use GPT-5 to extract inter-object relations. We first compute the bounding box trajectory for each object from its segmentation masks. To balance the relationship annotation density and computational cost, we sample the video at 1 FPS. For each sampled frame, we provide GPT-5 with the timestamp, the frame's height and width, and a list of visible objects along with their IDs and bounding boxes. The sampled frames and their corresponding metadata are fed to GPT-5 in an alternating sequence.

In practice, we observe that prompting GPT-5 to output all relation types in a single pass biases the model heavily toward spatial relations, causing many informative non-spatial relations to be missed. To mitigate this, we perform relation extraction in two separate passes using two dedicated prompts (Tables~\ref{tab:prompt_spatial} and \ref{tab:prompt_temporal})—one for spatial relations and one for non-spatial relations. For spatial relations, we prohibit trivial relations such as \textit{left of} or \textit{right of}, which can be determined directly from bounding boxes, and instead encourage the model to focus on relations that require genuine visual cues. For non-spatial relations, we designate $ID = -1$ as the video recorder (observer), enabling the model to extract salient relations between the camera and the objects. This two-pass strategy produces a more diverse and comprehensive set of relations for our dataset.

\noindent{\textbf{API Usage.}} We use \texttt{gpt-5-2025-08-07} Batch API for relation extraction, processing $\sim14B$ input tokens and generating $\sim5B$ output tokens.

\section{SVG2 Dataset Details}
\subsection{Data Statistics} \label{appdx_stats}
The SVG2 video samples are sourced from the SA-V dataset (43K) and the PVD dataset (593K). The synthesized annotations cover approximately 6.6M instance-level labels and trajectories, 52M attributes, and about 6.7M relationships.

\begin{figure}[!h]
    \centering
    \includegraphics[width=0.95\linewidth]{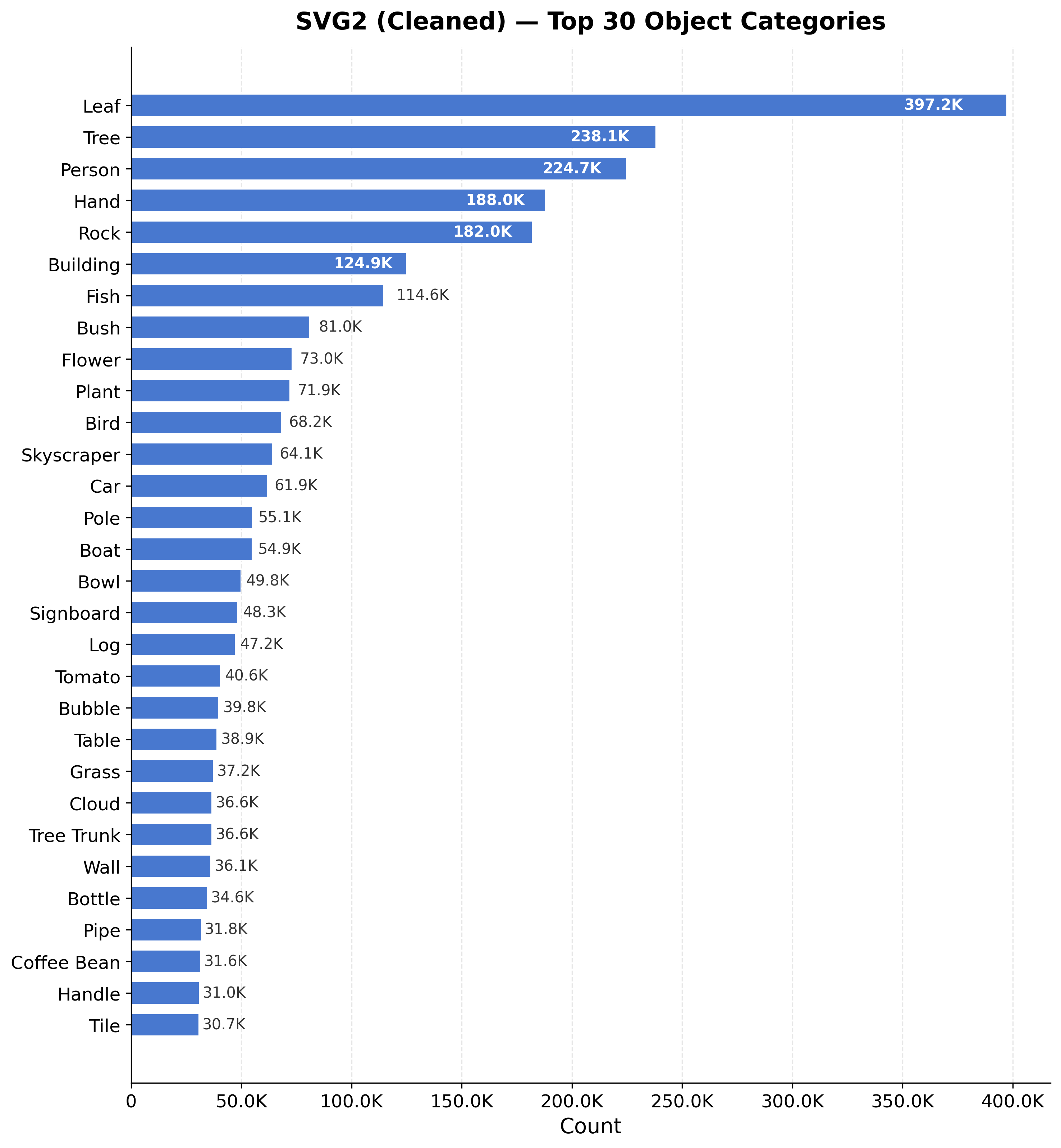}
    \caption{Distribution of object categories in SVG2. The dataset covers a diverse range of semantic classes including persons, vehicles, animals, furniture, and various everyday objects.}
    \label{fig:object_statistics}
\end{figure}

\begin{figure}[!h]
    \centering
    \includegraphics[width=0.95\linewidth]{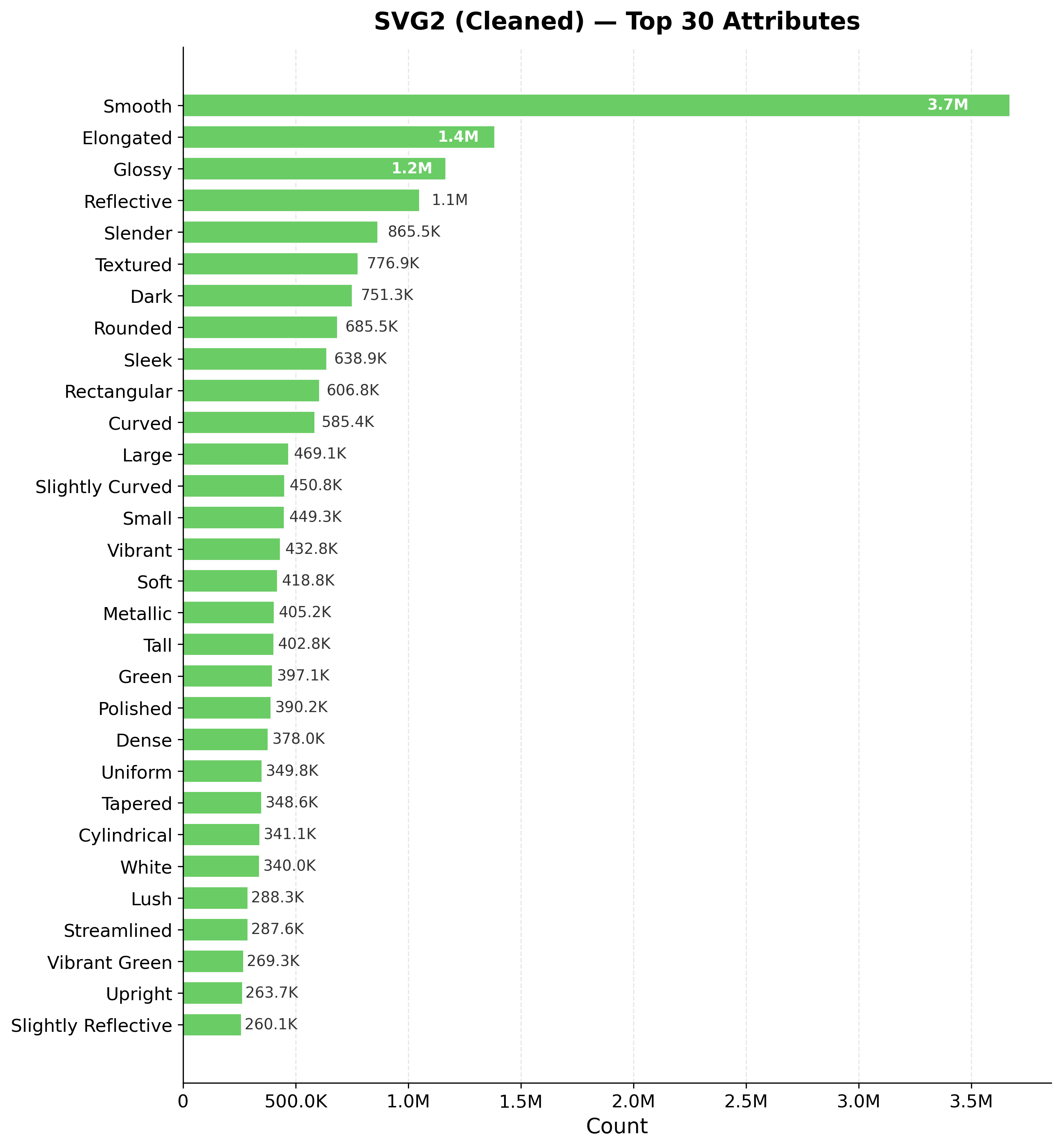}
    \caption{Distribution of attribute annotations in SVG2. The attributes span visual properties such as color, material, state, and other fine-grained descriptors.}
    \label{fig:attribute_statistics}
\end{figure}

\begin{figure}[!h]
    \centering
    \includegraphics[width=0.95\linewidth]{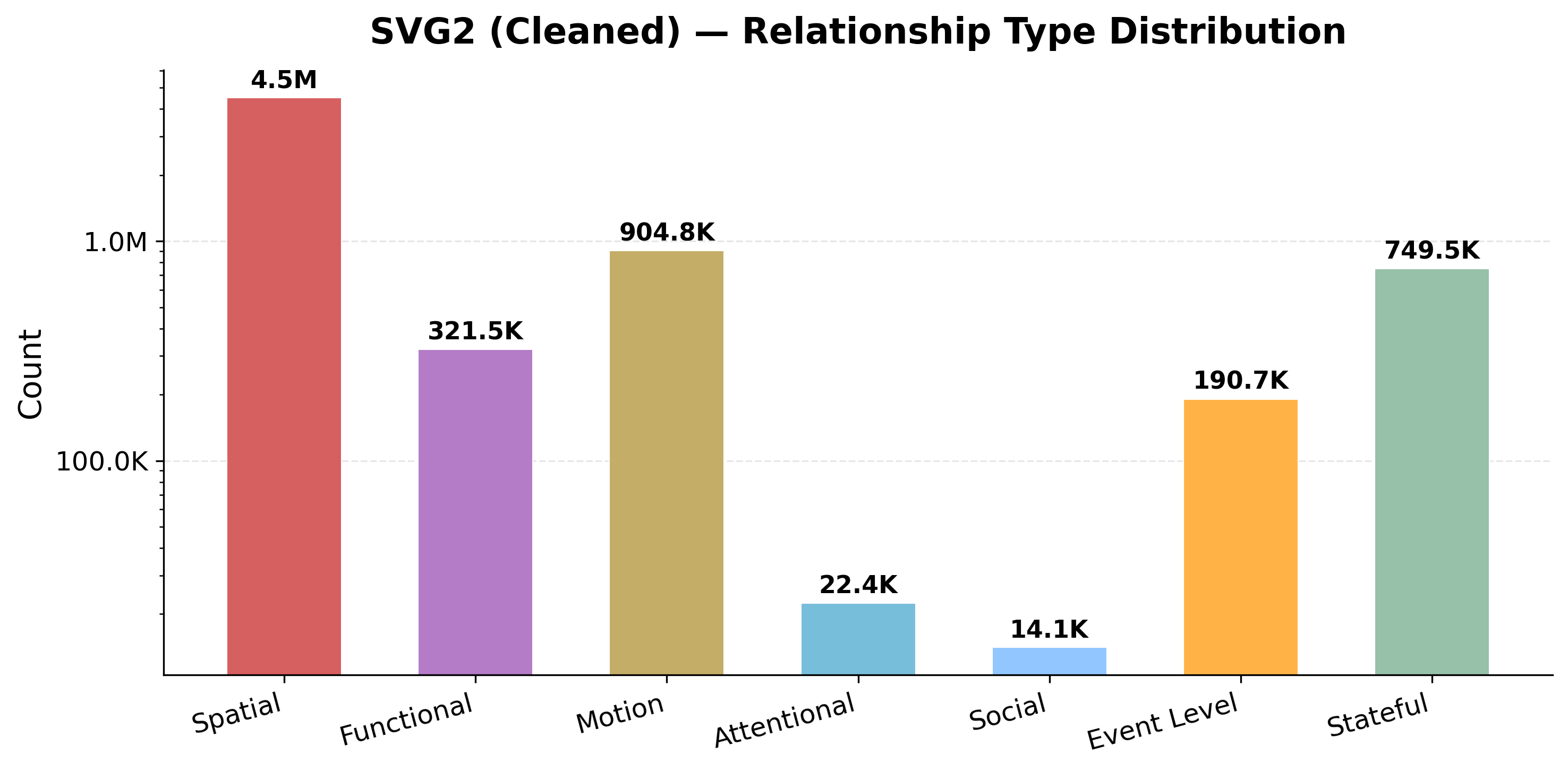}
    \caption{Overview of relationship categories in SVG2, showing the distribution across spatial, motion, functional, stateful, social, attentional, and event-level relation types.}
    \label{fig:rel_overview}
\end{figure}

\begin{figure}[]
    \centering
    \begin{subfigure}{0.3\textwidth}
        \includegraphics[width=\textwidth]{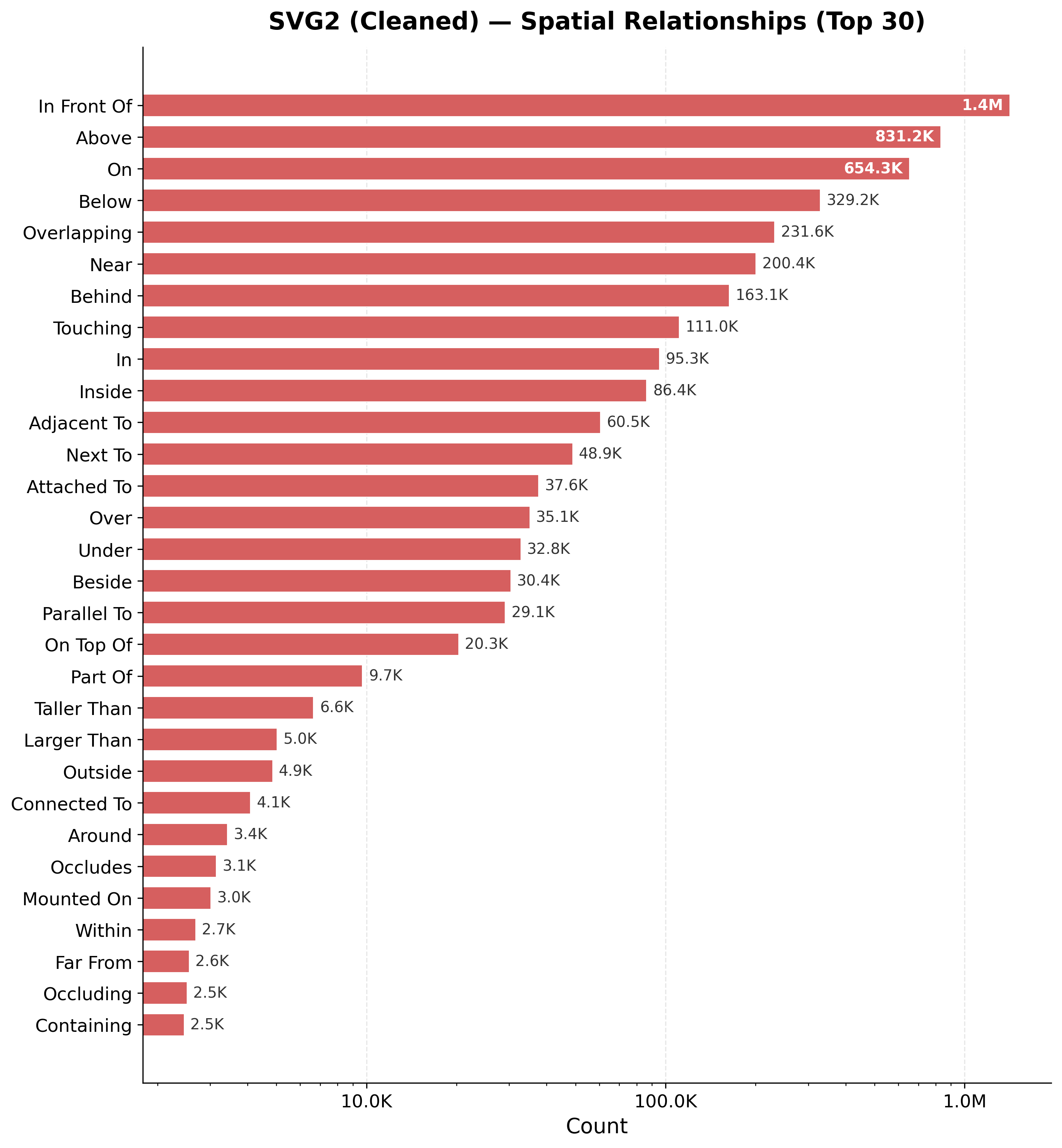}
        \caption{Spatial relations}
    \end{subfigure}
    \begin{subfigure}{0.3\textwidth}
        \includegraphics[width=\textwidth]{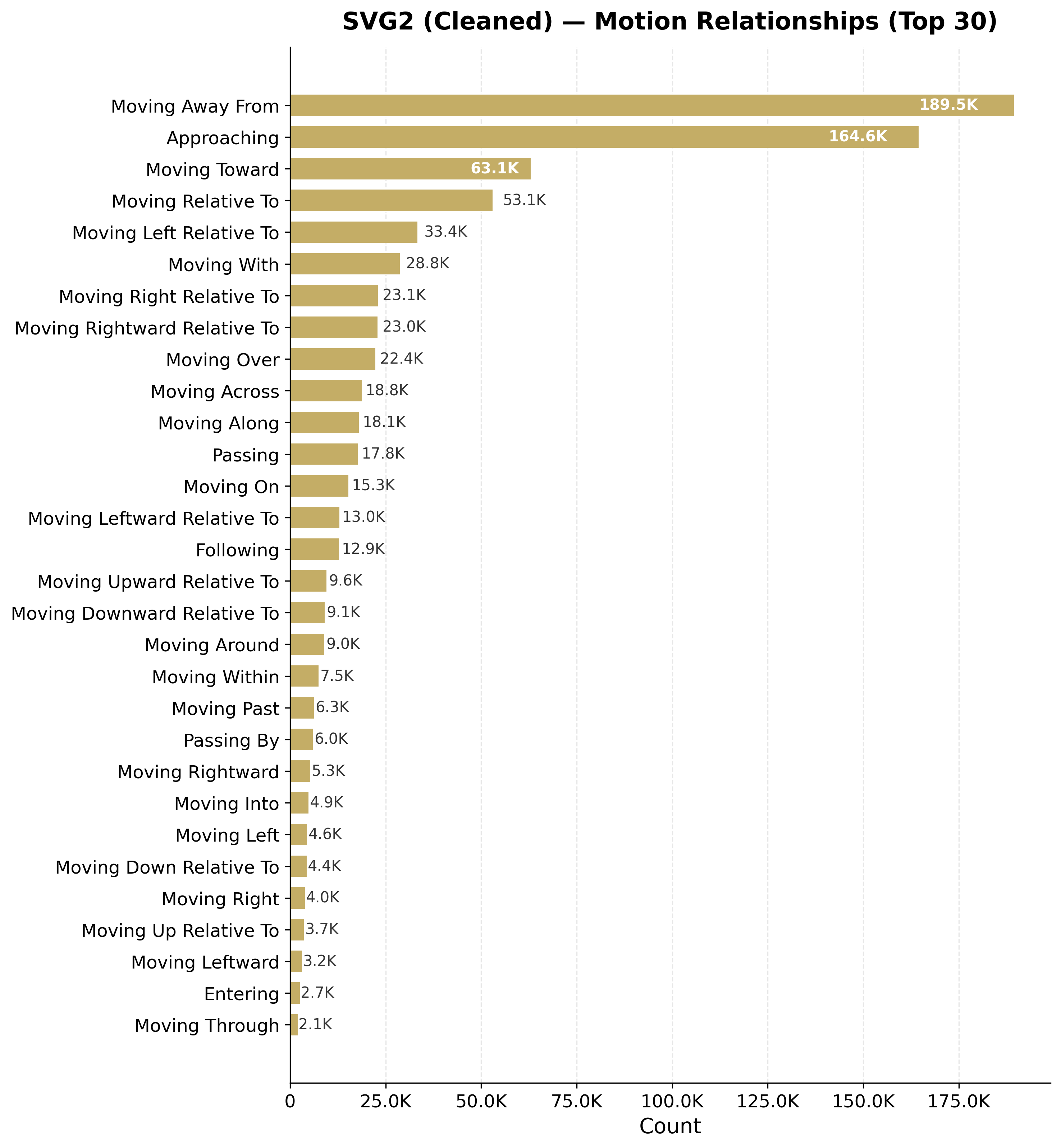}
        \caption{Motion relations}
    \end{subfigure}
    \begin{subfigure}{0.3\textwidth}
        \includegraphics[width=\textwidth]{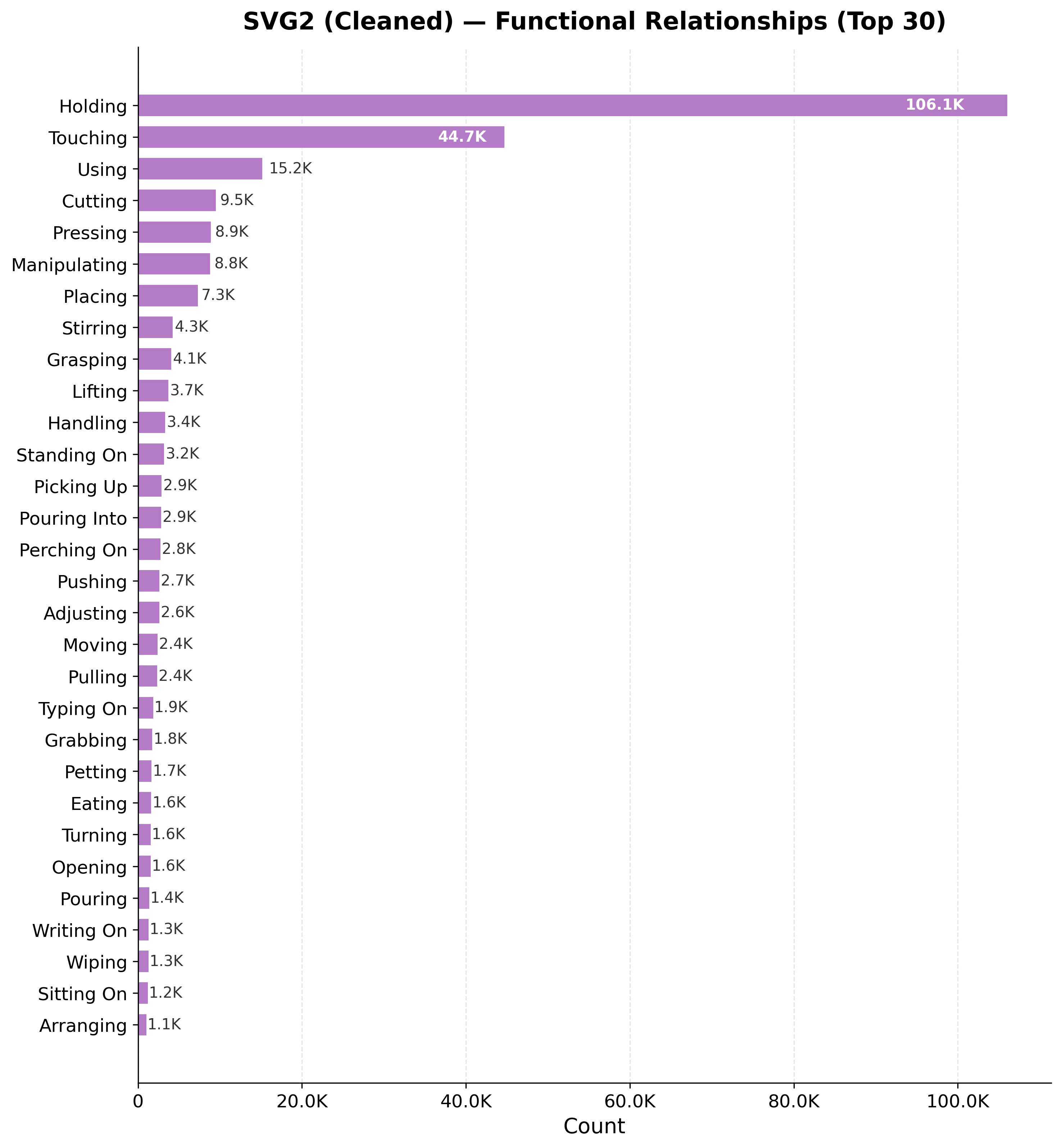}
        \caption{Functional relations}
    \end{subfigure}
    \begin{subfigure}{0.3\textwidth}
        \includegraphics[width=\textwidth]{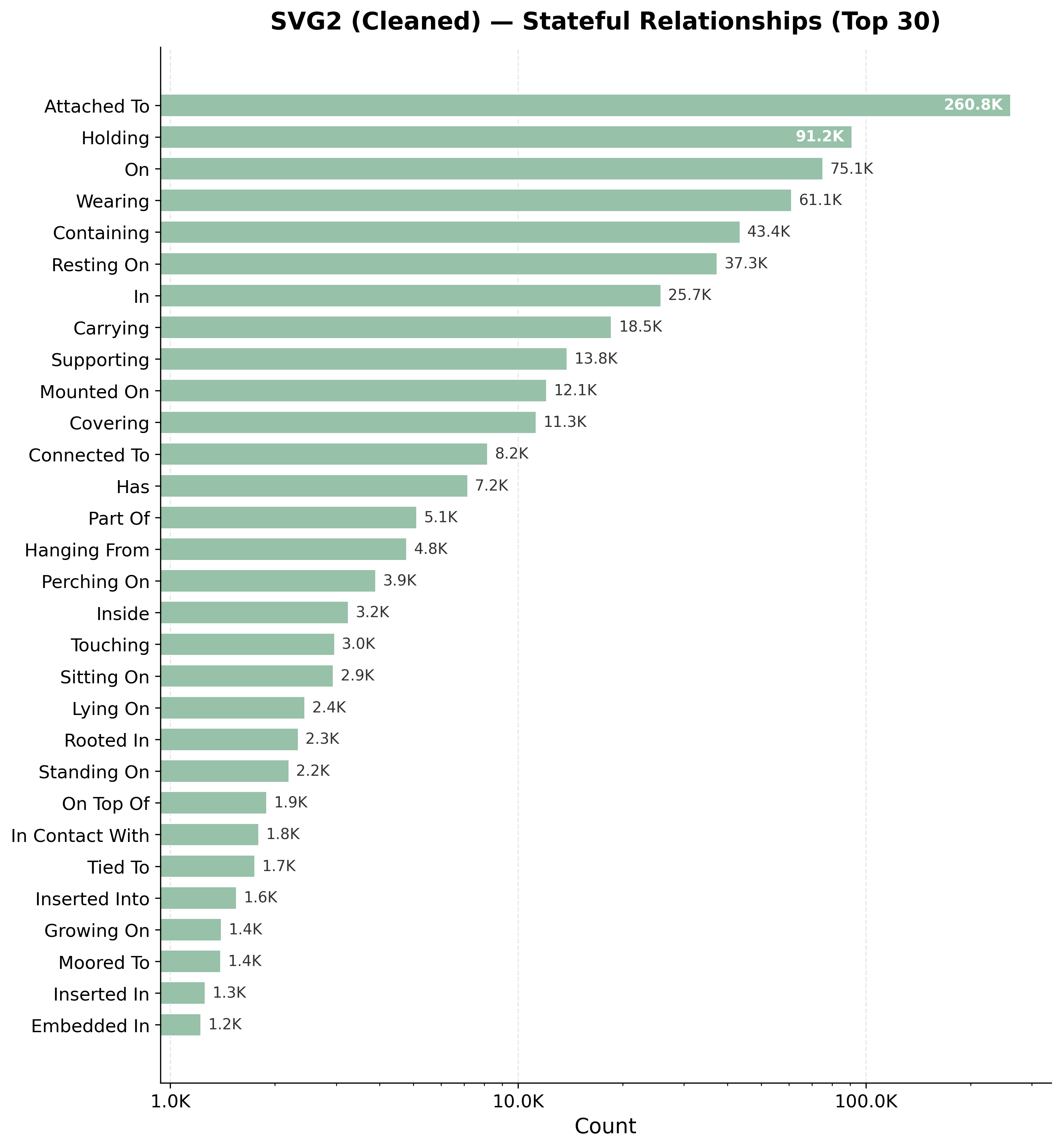}
        \caption{Stateful relations}
    \end{subfigure}
    \begin{subfigure}{0.3\textwidth}
        \includegraphics[width=\textwidth]{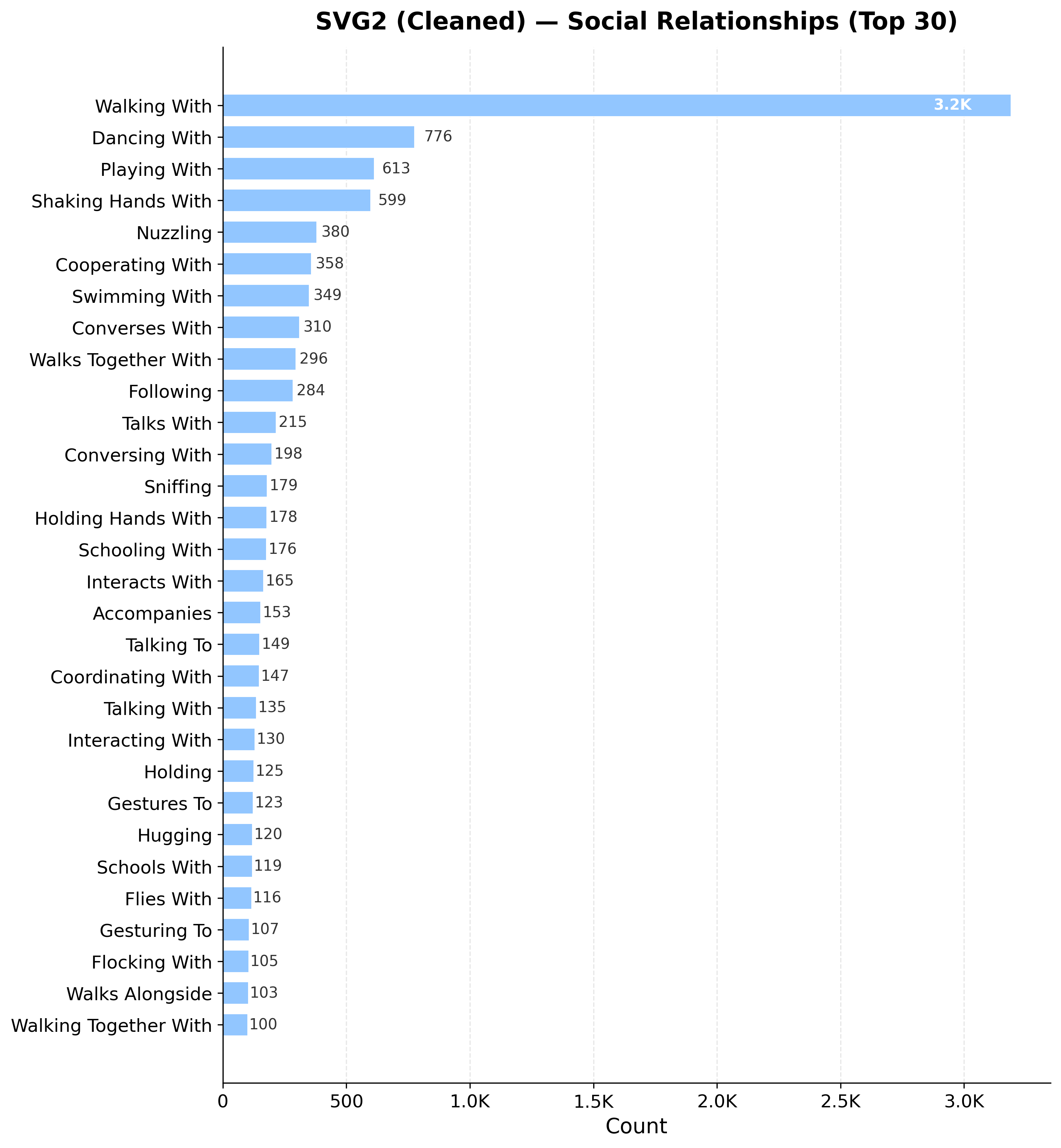}
        \caption{Social relations}
    \end{subfigure}
    \begin{subfigure}{0.3\textwidth}
        \includegraphics[width=\textwidth]{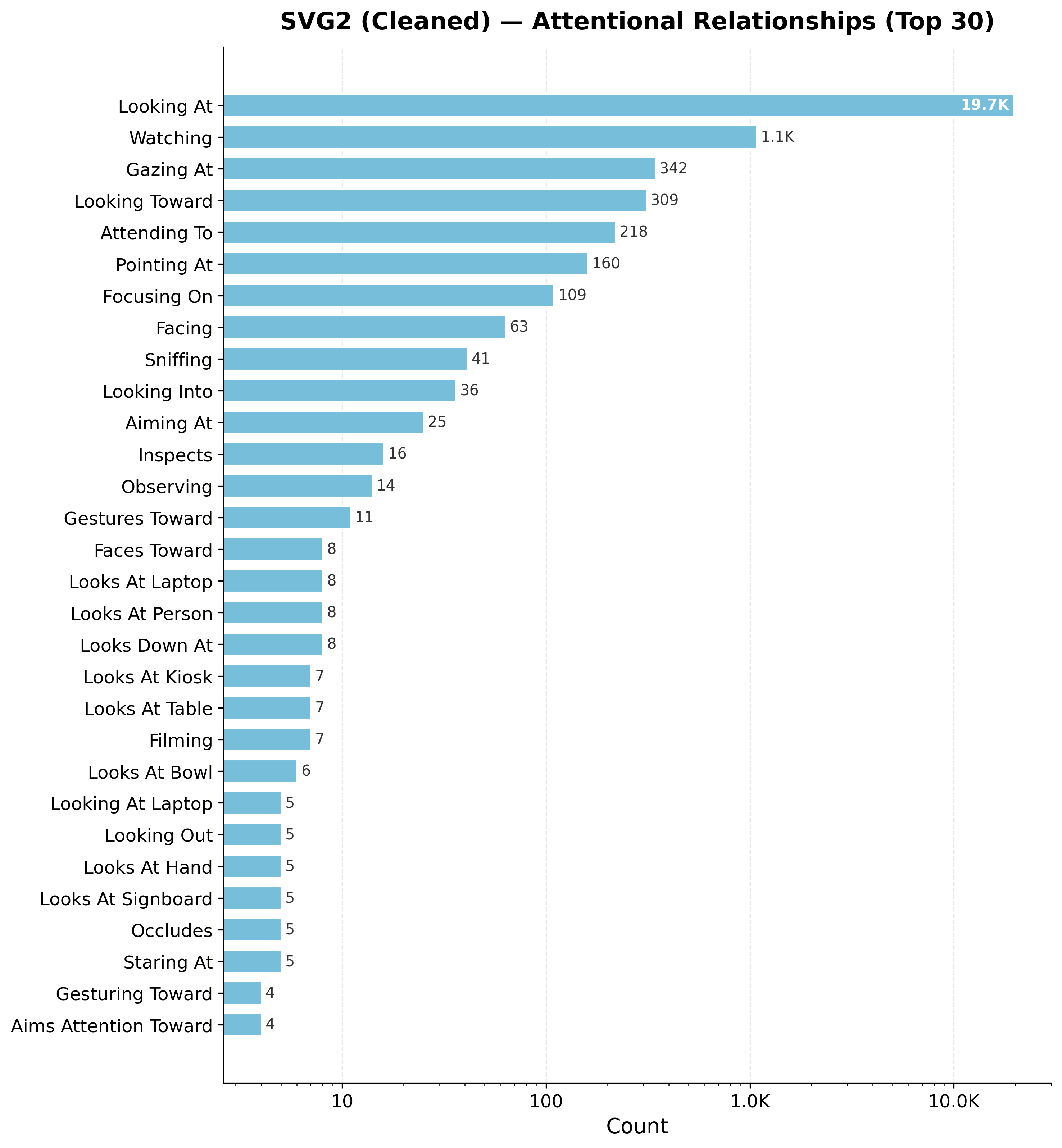}
        \caption{Attentional relations}
    \end{subfigure}
    \begin{subfigure}{0.3\textwidth}
        \includegraphics[width=\textwidth]{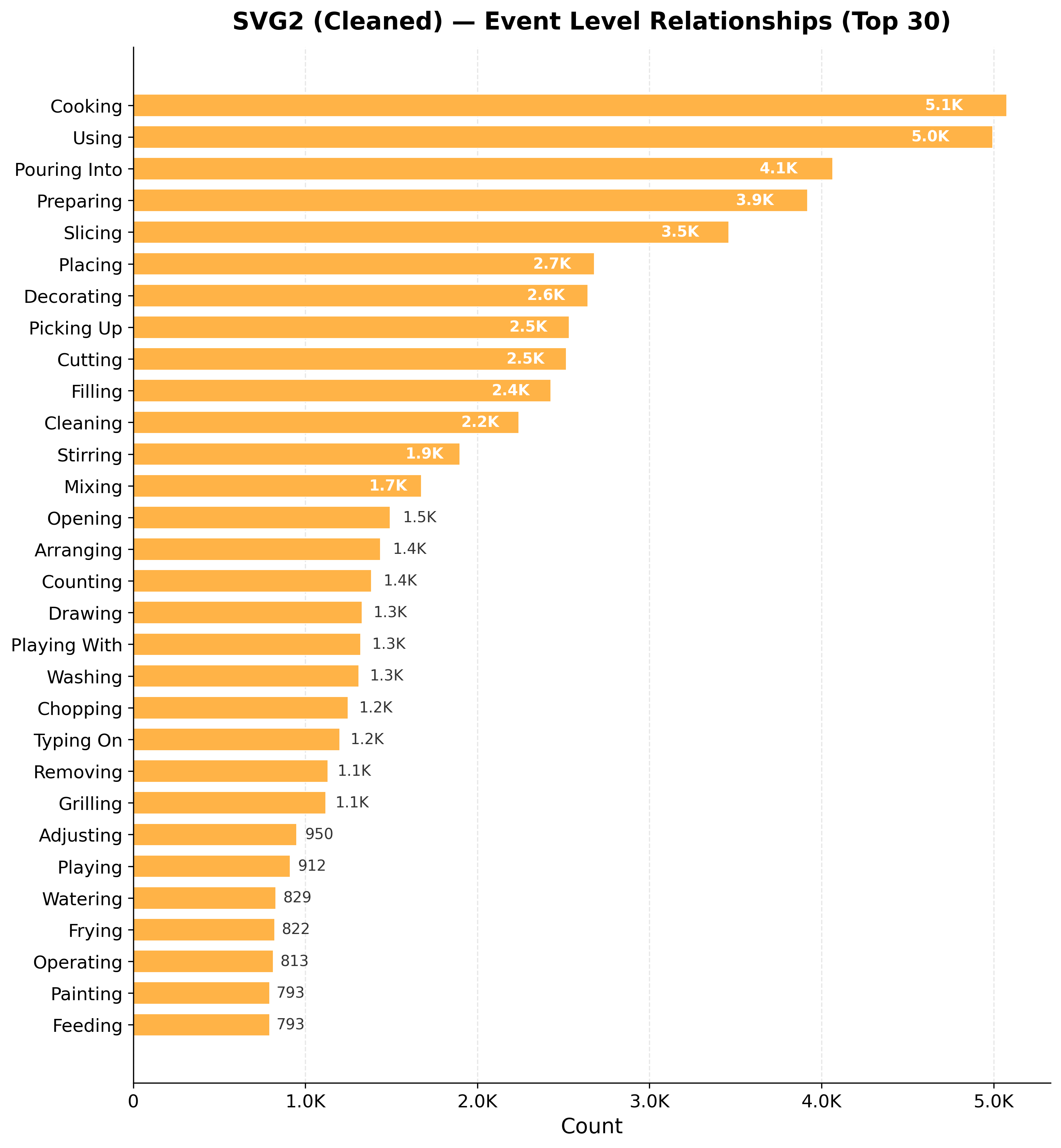}
        \caption{Event-level relations}
    \end{subfigure}

    \caption{Detailed distribution of relationship predicates across all seven categories in SVG2.}
    \label{fig:rel_subcategories}
\end{figure}

Figure~\ref{fig:object_statistics} shows the distribution of object categories in SVG2, illustrating the diversity and coverage of our dataset across various semantic classes.
Figure~\ref{fig:attribute_statistics} presents the distribution of attribute annotations, demonstrating the richness of fine-grained visual properties captured by our pipeline.
Figure~\ref{fig:rel_overview} provides an overview of relationship categories in SVG2, and Figure~\ref{fig:rel_subcategories} further shows the detailed distribution for each of the seven relationship types: spatial, motion, functional, stateful, social, attentional, and event-level. \Cref{tab:video_duration_stats} details the average video duration and frame rate statistics.

\begin{table}[]
\centering
\caption{Video duration and frame rate (FPS) statistics.}
\label{tab:video_duration_stats}
\scriptsize
\setlength{\tabcolsep}{6pt}
\renewcommand{\arraystretch}{1.}
\begin{tabular}{lccc}
\hline
\textbf{Dataset} &
\textbf{Duration (s)} &
\textbf{FPS} \\
\hline
SVG2-SAV      & $13.8 \pm 3.3$ & $24.0 \pm 0.0$ \\
SVG2-PVD       & $16.6 \pm 9.7$ & $29.6 \pm 8.1$ \\
SVG2 & $16.4 \pm 9.4$ & $29.2 \pm 8.0$  \\
$SVG2_{test}$ & $13.6 \pm 6.3$ & $11.8 \pm 8.4$  \\
\hline
\end{tabular}
\end{table}

\subsection{Visualizations}
\begin{figure}[!h]
    \centering
    \includegraphics[width=0.8\linewidth]{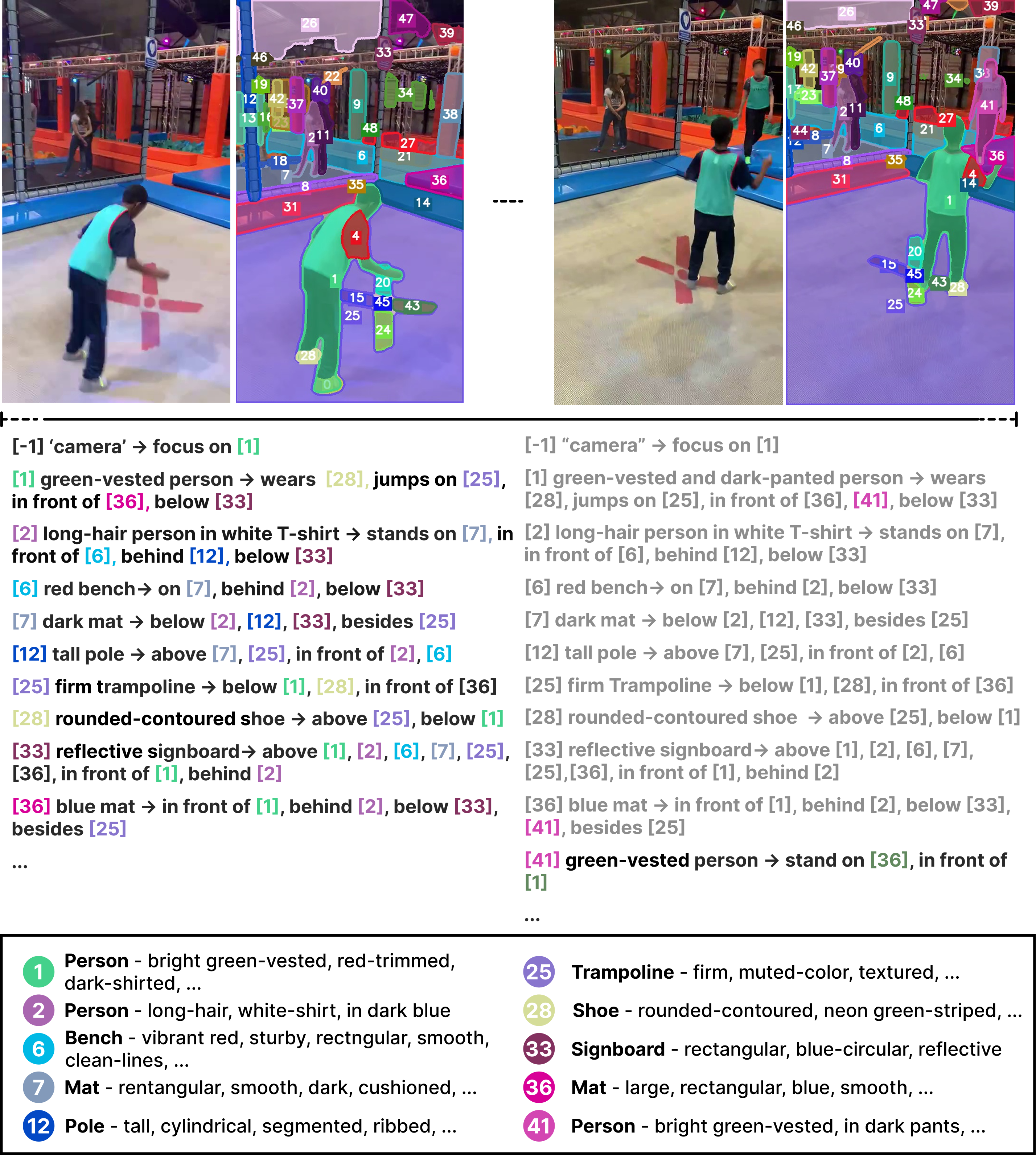}
    \caption{Example 1: Visualizations from the SVG2 dataset, illustrating panoptic trajectories, objects, attributes, and diverse inter-object relations.}

    \label{fig:vis_1}
\end{figure}

\begin{figure}[!h]
    \centering
    \includegraphics[width=0.8\linewidth]{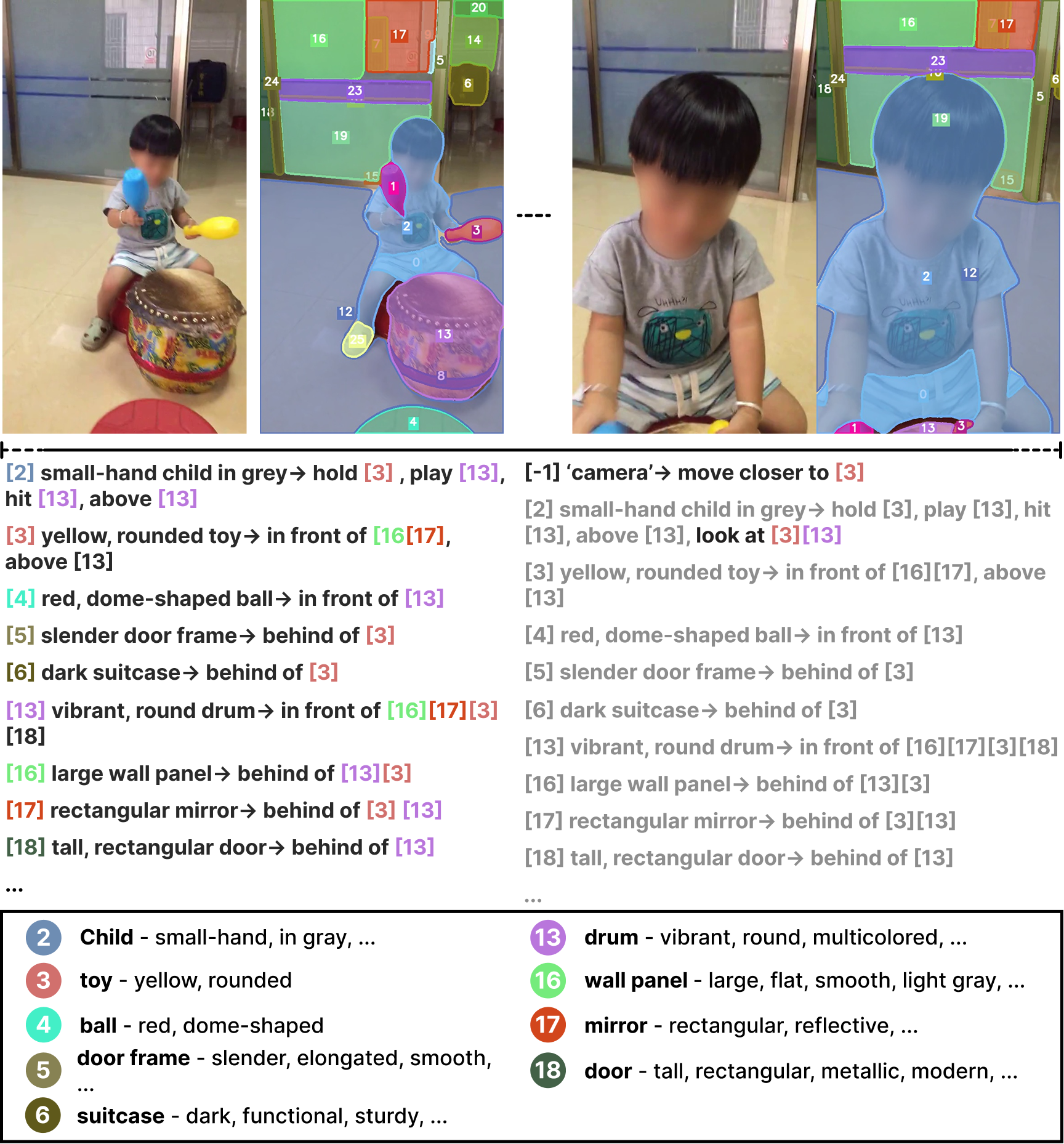}
    \caption{Example 2: Visualizations from the SVG2 dataset, illustrating panoptic trajectories, objects, attributes, and diverse inter-object relations.}

    \label{fig:vis_2}
\end{figure}

\begin{figure}[!h]
    \centering
    \includegraphics[width=0.8\linewidth]{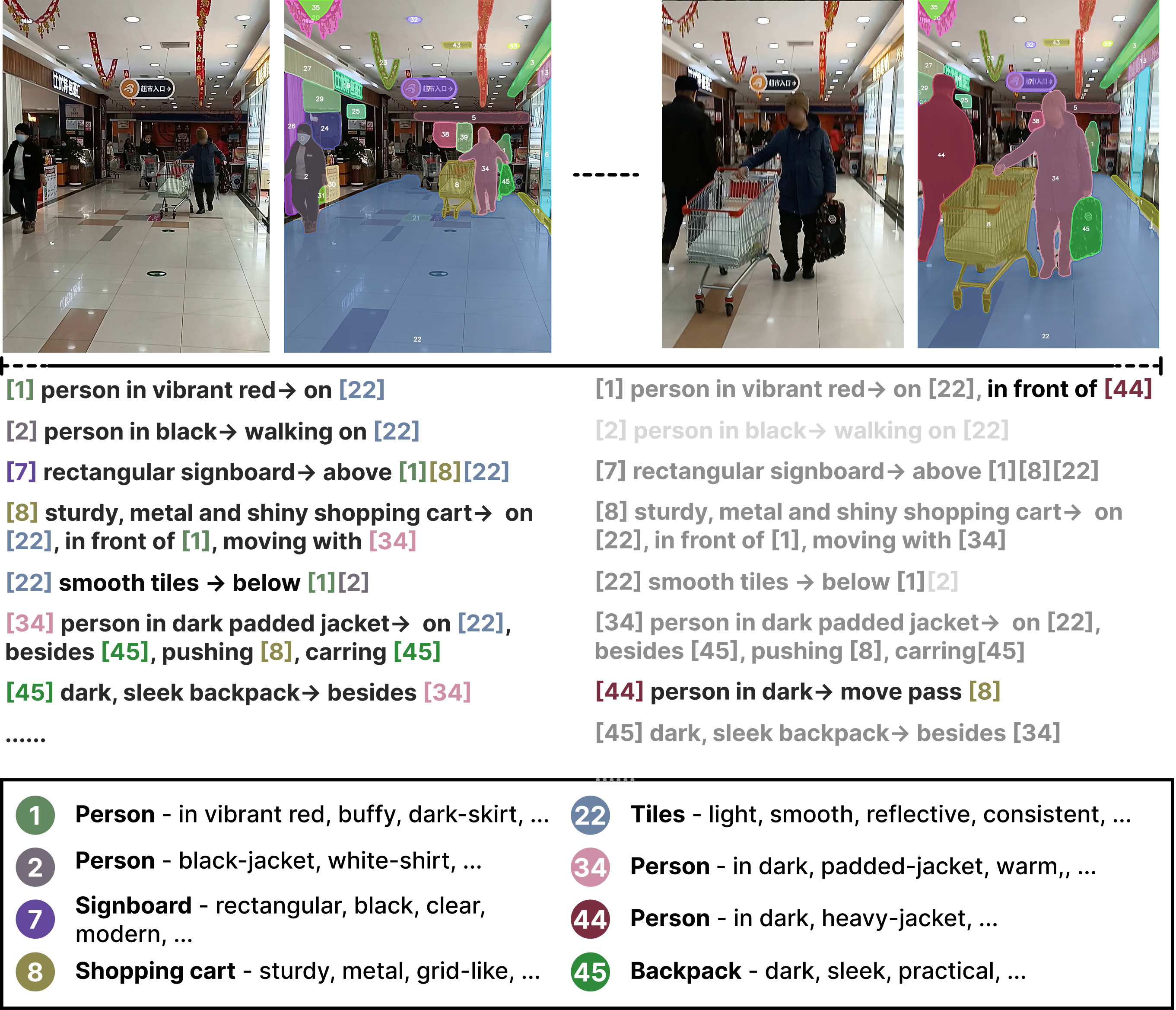}
    \caption{Example 3: Visualizations from the SVG2 dataset, illustrating panoptic trajectories, objects, attributes, and diverse inter-object relations.}

    \label{fig:vis_3}
\end{figure}

To further illustrate the structure of the SVG2 dataset, we present three annotation visualization examples, see Figures~\ref{fig:vis_1}, \ref{fig:vis_2}, and \ref{fig:vis_3}. These visualizations illustrate the panoptic object trajectories generated in Phase 1, fine-grained object semantics extracted from DAM and parsed into structured attributes, and inter-object relations derived from GPT-5 under both spatial and non-spatial prompting strategies.


\subsection{Human Annotation Workflow}

We rely on expert human annotators to building our $SVG2_{test}$ benchmark.
We curate expert annotators to create dense, panoptic video segmentation masks with scene graph annotations. We use the SAM2-based~\cite{sam2} annotation app to create hierarchical segmentation masks, where the first hierarchy captures objects as a whole and finer-grained levels annotate subparts for objects of interest. We recruit annotators within our institution who have been trained to use the SAM-2 editing framework to refine the segmentation masks, producing object labels and dense segmentation tracking at different levels of granularity.

\begin{figure}
    \centering
    \includegraphics[width=0.9\textwidth]{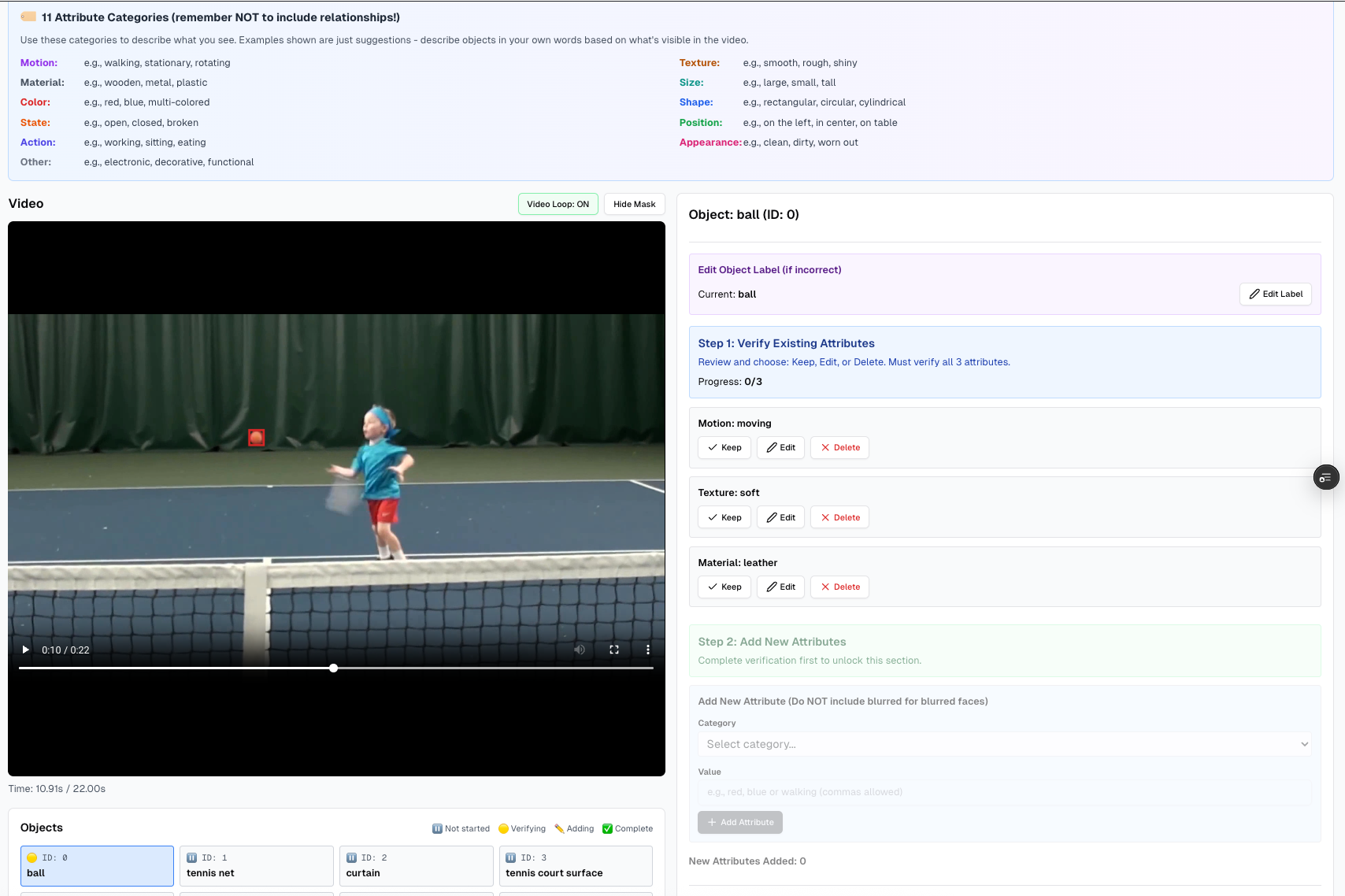}
    \caption{Crowdworkers editing and annotating objects and attributes from segmentation tracking. }
    \label{fig:annotation_ui_object_attributes}
    \centering
    \includegraphics[width=0.95\textwidth]{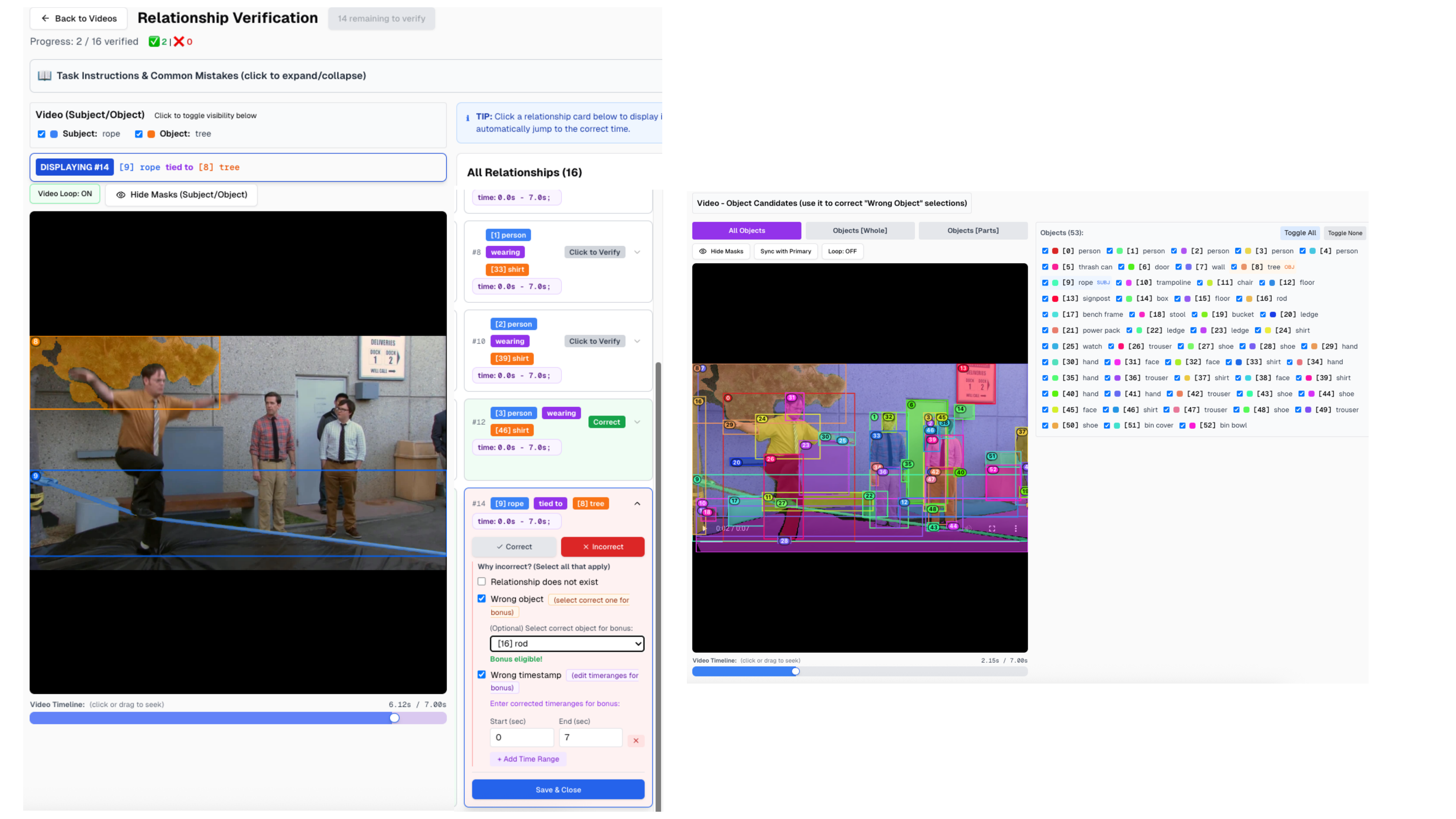}
    \caption{Crowdworkers verifying and fixing relationships. }
    \label{fig:annotation_ui_rel}
\end{figure}

For scene graph annotations, we employ a model-in-the-loop approach. We prompt GPT-5 with human-annotated object labels to generate initial attributes and relationships with timestamps, dividing relationships into spatial and temporal categories. We then use the Prolific\footnote{https://www.prolific.com/} crowd-sourcing platform to recruit workers who have been identified as experts in video annotation. Annotators review and edit object labels and attributes, choosing to keep, remove, add, or modify each entry. They are required to annotate at least one attribute per object and are compensated based on the number of attributes annotated. For relationships, annotators perform similar edits, removing incorrect relationship labels, adding new relationship labels, fixing mismatched object IDs, or adjusting timestamps. When object IDs or timestamps are incorrect, annotators can select from all available objects in the video or modify the temporal range where the relationship applies. Figures~\ref{fig:annotation_ui_object_attributes}--\ref{fig:annotation_ui_rel} show the annotation interfaces used for verification and editing.


\section{Model Details}
\begin{figure}[!h]
    \centering
    \includegraphics[width=0.6\linewidth]{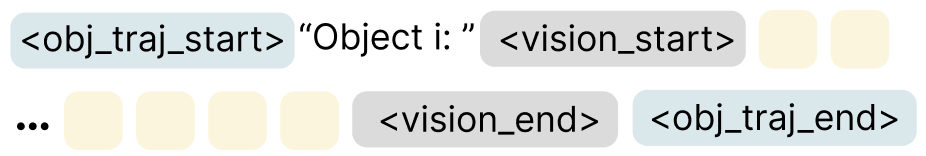}
    \caption{Rearranged object-level token sequence example. 
We introduce two new special tokens, \textit{\texttt{<obj\_traj\_start>}} and \textit{\texttt{<obj\_traj\_end>}}, to delimit individual trajectories. 
Within each trajectory, we explicitly mark the object identity using the text \textit{``Object i''}, and we continue to use the existing special tokens from the Qwen2.5-VL-3B vocabulary to separate visual tokens.}

    \label{fig:token_seq}
\end{figure}

Our proposed \ourmodel introduces a trajectory-aligned token arrangement mechanism, designed to map valid tokens according to the pixel regions covered by each object's trajectory. However, the rearrangement disrupts the original order and spatial layout of the visual tokens. Therefore, for the token sequence associated with each object's trajectory, we normalize it into the structured format illustrated in Figure~\ref{fig:token_seq}. Specifically, we prepend a pair of special tokens, \texttt{<obj\_traj\_start>} and \texttt{<obj\_traj\_end>}, to explicitly delineate different trajectories. Within each trajectory segment, we insert the textual identifier of the object (i.e., its corresponding text embedding), which matches the object ID used in the video-scene-graph supervision signals. We then continue to leverage the pretrained special tokens from Qwen2.5-VL to separate the visual tokens associated with each object.

During the compression stage via the dual-resampler, the tokens of each object, either as a whole or partitioned into temporal windows, are fed into two resamplers, while the special tokens and object-ID embeddings bypass the resamplers. The output tokens from the object-trajectory resampler are placed first, followed by the tokens produced by the temporal-window resampler for each respective window.

\section{Training Details}
During training, we set the token-arrangement selection threshold $\tau_{\mathrm{eff}}$ to 0.5 to balance contextual coverage and token count. Both resamplers are configured with three layers and 32 learnable queries. We apply RMSNorm both before and after the cross-attention blocks in each layer. The hidden size of each resampler is set to 2048, and the intermediate size of each MLP layer is $4\times2048$.

We sample video frames at 1~fps and adopt the same resizing strategy as Qwen2.5-VL. The ViT backbone is frozen, while the projector, language model, and the newly introduced resamplers are trained. For the temporal-window resampler, each temporal window spans 4~s, meaning that the token groups corresponding to different temporal segments of a trajectory are processed independently. The object-trajectory resampler processes all tokens belonging to each object in a single pass.

We use a warmup ratio of 0.03 and a weight decay of 0.01. Training is performed on 8 H100 GPUs with a batch size of 1 and gradient accumulation steps set to 2.

\section{Additional Experiments and Results}\label{appx:more_results}
\subsection{Evaluation details}
To ensure a fair comparison across all baseline models, we provide each model with the bounding-box trajectory of every object during evaluation. For each bounding box, we include both its absolute coordinates and normalized relative coordinates, and we explicitly prompt the model with the original frame resolution.
For models that support direct video input, for example, Gemini 2.5 Pro, we supply the bounding boxes and their corresponding timestamps computed from frames sampled at 1 fps. For models that do not support video input, such as GPT-4.1 and GPT-5, we instead feed the 1 fps (which is the same as \ourmodel inference setting) sampled frames sequentially, alternating each frame with its associated metadata and the bounding-box information of all objects in that frame.

All models use prompts following a unified structural template, and their outputs are constrained using a predefined JSON schema.


\subsection{Comparison under different evaluation criteria}

\begin{table}[t]
\centering
\caption{Comparison results under different evaluation criteria. We evaluate models using a relaxed temporal grounding threshold for relations and triplets (IoU 0.1). For object evaluation, we report a rule-based 'Strict' score, which requires an exact string match between the predicted and ground-truth labels.}
\scriptsize
\setlength{\tabcolsep}{2pt}
\renewcommand{\arraystretch}{0.9}

\resizebox{\textwidth}{!}{
\begin{tabular}{l ccc ccc cccc }
\toprule
& \multicolumn{3}{c}{\textbf{Triplet (IoU 0.1)}}
& \multicolumn{3}{c}{\textbf{Relation (IoU 0.1)}}
& \multicolumn{4}{c}{\textbf{Object (Strict)}}\\

\cmidrule(lr){2-4} \cmidrule(lr){5-7} \cmidrule(lr){8-11} 
\textbf{Model}
& PVSG & VidOR & SVG2\textsubscript{test}
& PVSG & VidOR & SVG2\textsubscript{test}
& VIPSeg & PVSG & VidOR & SVG2\textsubscript{test}
\\
\midrule
\multicolumn{11}{l}{\textbf{\emph{API call only}}} \\
GPT-4.1~\cite{achiam2023gpt}
& 10.3 & 15.4 & 6.8
& 12.3 & 17.2 & 8.1
& 30.0 & 15.0 & \underline{26.9 }& 24.9
\\
Gemini-2.5 PRO~\cite{comanici2025gemini25}
& \underline{12.3} & 12.8 & 9.3
& {13.9} & 14.3 & 10.6
& 28.0 & 9.2 & 20.7 & 23.3
\\
GPT-5~\cite{gpt5}
& \textbf{23.8} & \underline{25.6} & \textbf{18.9}
& \underline{24.2} & \underline{28.5} & \underline{20.5}
& \underline{38.2} & \underline{20.1} & {22.2} & \textbf{33.4}
\\
\midrule
\multicolumn{11}{l}{\textbf{\emph{Open weights only}}} \\
Qwen2.5-VL-3B~\cite{bai2025qwen25vl}
& 0.1 & 0.4 & 0.3
& 0.1 & 0.8 & 0.6
& 15.6 & 2.4 & 9.5 & 9.1
\\
MiniCPM-V 4.5~\cite{yao2024minicpmv}
& 0.2 & 5.0 & 1.3
& 0.5 & 6.6 & 2.7
& 25.5 & 0.5 & 12.0 & 16.4
\\
InternVL3.5-4B~\cite{wang2025internvl35}
& 0.6 & 2.1 & 1.0
& 1.8 & 2.9 & 2.1
& 20.9 & 2.9 & 10.6 & 14.7
\\
GLM-4.1-9B-Thinking~\cite{zeng2024chatglm}
& 0.7 & 5.7 & 2.1
& 1.5 & 7.2 & 3.5
& 23.2 & 2.6 & 10.4 & 11.7
\\
Qwen3-VL-4B~\cite{yang2025qwen3}
& 0.1 & 1.1 & 1.1
& 0.1 & 1.4 & 1.9
& 22.8 & 4.4 & 14.2 & 16.7
\\
Qwen3-VL-4B-Thinking~\cite{yang2025qwen3}
& 0.1 & 3.7 & 2.0
& 1.5 & 5.7 & 4.2
& 22.9 & 0.8 & 10.7 & 16.6
\\
\midrule
\multicolumn{11}{l}{\textbf{\emph{Fine-tuned baselines}}} \\
FT-Qwen2.5-VL-3B (First Bbox)
& 0.1 & 3.5 & 1.2
& 0.9 & 10.0 & 1.9
& 11.5 & 5.9 & 11.7 & 9.8 \\
FT-Qwen2.5-VL-3B (Bbox Traj.)
& 2.5 & 3.5 & 1.7
& 6.6 & 8.4 & 3.8
& 11.0 & 7.5 & 13.7 & 13.1\\
\midrule
\ourmodel
& \textbf{23.8} & \textbf{32.7} & \underline{18.5}
& \textbf{25.4} & \textbf{36.0} & \textbf{20.8}
& \textbf{42.1} & \textbf{29.5} & \textbf{31.7} & \underline{28.8}
\\
\bottomrule
\end{tabular}
}
\label{tab:strictbase}
\end{table}

We report additional results under the alternative evaluation criteria introduced in~\cref{sec:experiment}. \Cref{tab:strictbase} presents triplet and relation recall using a relaxed temporal IoU threshold of 0.1. For object evaluation, we report the rule-based \textbf{strict score}, which requires exact string matches between the predicted and ground-truth labels.

As shown in the table, proprietary models perform reasonably well across all tasks, with GPT-5 achieving the strongest overall results among API-access models. In contrast, open-weight VLMs display a substantial performance gap, particularly on triplet and relation prediction. This underscores the inherent difficulty of fine-grained, temporally grounded scene graph extraction for current open-source models.

\ourmodel, however, demonstrates clear and consistent improvements across all benchmarks. Under the relaxed IoU criterion, it achieves strong triplet and relation performance, surpassing all open-weight baselines by large margins. Crucially, under the exact-match \textbf{strict score}, \ourmodel continues to deliver the highest object-level performance across all datasets, frequently exceeding even the proprietary API models. These results confirm that our trajectory-aligned token arrangement and dual-resampler design significantly enhance fine-grained object grounding and temporal consistency, both of which are critical for accurate video scene graph generation.

\subsection{Additional Ablations}
\paragraph{Model architecture.} We further investigate the impact of the resampler architecture by varying the number of resampler layers and the number of learnable queries in the temporal-window resampler. All variants are trained for 3 epochs on the SAV subset of SVG2. As shown in Table~\ref{tab:model-arch}, the performance differences across architectural variants are relatively small. The configuration with three layers and 32 queries achieves slightly better overall results and inference stability.

\begin{table}[!h]
\centering
\caption{Data scale ablation showing consistent average performance gains as training data increases.}
\label{tab:datascale}
\scriptsize 
\setlength{\tabcolsep}{6pt} 
\renewcommand{\arraystretch}{1.1} 
\begin{tabular}{lcccc}
\toprule
Metric / Data scale & 25\% & 50\% & 75\% &  100\% \\
\midrule
\multicolumn{4}{l}{\textbf{Triplet}} \\
\quad PVSG  &  10.1 &  12.2 &  13.7 & 16.1\\
\quad VidOR & 20.7 &   20.5  &   22.4     & 22.9 \\
\quad SVG2\textsubscript{test}  &  17.5 & 18.9 &  15.7 & 16.7\\
\midrule
\multicolumn{4}{l}{\textbf{Relation} } \\
\quad PVSG  & 10.8 & 13.1 &  15.4 & 16.9\\
\quad VidOR & 22.9 &   22.5   & 24.7 &  25.0 \\
\quad SVG2\textsubscript{test}  & 19.6 & 20.9 &  18.3 & 18.7 \\
\midrule
\multicolumn{4}{l}{\textbf{Object}} \\
\quad PVSG   & 76.3 & 76.6 &  75.5 & 72.7\\
\quad VidOR  & 90.8 & 91.2 & 91.1 & 91.4\\
\quad VIPSeg & 85.7 & 85.3 & 85.7 & 86.5\\
\quad SVG2\textsubscript{test}   & 79.0 & 79.0 &79.1 & 79.0\\
\midrule
\multicolumn{4}{l}{\textbf{Attribute}} \\
\quad SVG2\textsubscript{test}  & 22.5   & 22.3 & 22.9 & 27.1\\
\midrule
\textbf{Avg Rank} 
& 4 & 3 & 2 & 1 \\
\bottomrule
\end{tabular}
\end{table}

\paragraph{Training data scale.} We analyze the effect of training data scale by comparing models trained with 25\%, 50\%, 75\%, and 100\% of the data. As shown in Table~\ref{tab:datascale}, increasing the amount of training data consistently improves the average performance across most metrics and benchmarks. For triplet and relation prediction, both VidOR and SVG2\textsubscript{test} exhibit clear gains as data scale grows, confirming that learning temporal and relational structures benefits strongly from large SVG2 dataset. Attribute prediction on SVG2\textsubscript{test} shows the most pronounced improvement, with a substantial jump from 22.3 to 27.1 when using the full dataset. This indicates that our 593K PVD subset in SVG2 provides diverse and high-quality object-level annotations. The overall average rank demonstrates that larger-scale training yields consistently better performance across all evaluation dimensions.

\begin{table}[]
\centering
\caption{Ablation study on resampler depth and query count. We evaluate the impact of varying the number of layers (Depth) and the number of learnable latent queries (\#query) within the Temporal-Window Resampler (TWR). *PVSG results are evaluated on the VidOR subset.}
\label{tab:model-arch}
\scriptsize 
\setlength{\tabcolsep}{6pt} 
\renewcommand{\arraystretch}{1.1} 
\begin{tabular}{cccccccc}
\hline
\multicolumn{2}{c}{Variants} &
\multicolumn{2}{c}{\textbf{Triplet}} &
\multicolumn{2}{c}{\textbf{Relation}} &
\multicolumn{2}{c}{\textbf{Object}} \\
\cmidrule(lr){1-2} \cmidrule(lr){3-4} \cmidrule(lr){5-6} \cmidrule(lr){7-8}

Depth &
\# query of TWR &
PVSG* & VidOR & PVSG* & VidOR & PVSG* & VidOR \\
\hline

2 & 16 & 12.3 & 15.1 & 13.5 & \textbf{16.5 }& \textbf{81.3 }& 89.9 \\
3 & 16 & 12.8 & 14.2  & 11.7 & 15.4  & 80.1 & \textbf{90.2} \\
2 & 32 & 12.0 & \textbf{15.6} & 14.7 & 17.5 & 80.9 & 90.0 \\
3 & 32 & \textbf{13.6} & 15.0 & \textbf{15.6} & 16.3 & 79.7 & \textbf{90.2} \\
\hline
\end{tabular}
\end{table}

\subsection{Trajectory Robustness and Temporal Scalability}
\noindent{\textbf{Robustness to upstream segmentation.}}
To isolate scene-graph generation quality, our primary open-vocabulary VidSGG protocol evaluates models using ground-truth panoptic trajectories. However, real-world deployment relies on predicted upstream segmentation. To assess \ourmodel's sensitivity to trajectory quality, we replace the ground-truth inputs with masks generated entirely by our automated Phase-1 tracking pipeline, retaining predicted masks that achieve an IoU $\ge$ 0.3 against the ground truth.

Table~\ref{tab:e2e_compare} compares performance between these input settings. While using predicted trajectories naturally reduces overall metrics due to upstream segmentation noise and different segmentation granularity, \ourmodel remains remarkably robust. Even with noisy predicted masks, \ourmodel still outperforms GPT-4.1 and Gemini-2.5 Pro (which were provided perfect ground-truth inputs), and all the open-source models across triplet, relation, object, and attribute metrics. This demonstrates that \ourmodel's trajectory-aligned token arrangement and dual-resampler design effectively mitigate realistic segmentation imperfections, proving adaptable to fully automated pipelines.

\begin{table}[!htbp]
\centering
\caption{Sensitivity to upstream trajectory quality. We compare \ourmodel using ground-truth panoptic segmentation masks (used in all main experiments) versus pipeline-predicted masks matched at IoU $\ge$ 0.3. Even with noisy and multi-granularity predicted trajectories, \ourmodel maintains highly competitive performance. $^\dagger$PVSG is evaluated on its VidOR subset only.}
\label{tab:e2e_compare}
\scriptsize 
\setlength{\tabcolsep}{6pt} 
\renewcommand{\arraystretch}{1.1} 
\resizebox{\textwidth}{!}{
\begin{tabular}{l ccc ccc cccc c}
\toprule
& \multicolumn{3}{c}{\textbf{Triplet}}
& \multicolumn{3}{c}{\textbf{Relation}}
& \multicolumn{4}{c}{\textbf{Object}}
& \multicolumn{1}{c}{\textbf{Attribute}} \\
\cmidrule(lr){2-4} \cmidrule(lr){5-7} \cmidrule(lr){8-11} \cmidrule(lr){12-12}
\textbf{Input Trajectory}
& PVSG$^\dagger$ & VidOR & SVG2\textsubscript{test}
& PVSG$^\dagger$ & VidOR & SVG2\textsubscript{test}
& VIPSeg & PVSG$^\dagger$ & VidOR & SVG2\textsubscript{test}
& SVG2\textsubscript{test} \\
\midrule
pipeline-predicted 
& {10.0} & {4.3} & {10.6}
& {13.4} & {6.1} & {16.2}
& {72.2} & {63.4} & {62.9} & {64.1}
& {17.6} \\
ground-truth &
{19.2} & {22.9} & {16.7} &
{20.1} & {25.0} & {18.7} &
{86.5} & {81.9} & {91.4} & {79.0} &
{27.1} \\
\bottomrule
\end{tabular}
}
\end{table}

\noindent{\textbf{Temporal scalability and long-video generalization.}}
Our \ourdataset training videos average $\sim$15 seconds in duration. To evaluate zero-shot temporal generalization, we test \ourmodel on the PVSG benchmark's subsets, which span durations up to $\sim$10$\times$ longer: VidOR (avg. 48s), Epic-Kitchens (120s), and Ego4D (166s). In our experiments, \ourmodel processes these extended sequences directly without truncation, successfully handling up to 300 frames (at 1 fps) for the longest videos in the test set.
Table~\ref{tab:video_length_compare} details this zero-shot temporal generalization. Object prediction accuracy remains highly robust across all durations, demonstrating that the object-trajectory resampler's global aggregation extracts semantically stable identity representations regardless of video length. Relation and triplet performance exhibits a decline, reflecting the inherent difficulty of extrapolating fine-grained temporal grounding to horizons far beyond the training distribution. One potential strategy to better balance long-form video support is to reduce the number of learnable queries within the temporal-window resampler. 
These results highlight the generalization capability of our model while identifying long-video VSG as a promising direction for future work: explicitly incorporating longer training videos and scalable temporal architectures could substantially close the performance gap at extended durations.

\begin{table}[!htbp]
\centering
\caption{Zero-shot temporal generalization across three PVSG subsets spanning durations from 48s (VidOR) to 166s (Ego4D). Object prediction remains highly robust across all lengths, while relation and triplet performance decreases for longer videos, motivating future long-video training strategies.}
\label{tab:video_length_compare}
\scriptsize 
\setlength{\tabcolsep}{6pt} 
\renewcommand{\arraystretch}{1.1} 
\resizebox{\textwidth}{!}{
\begin{tabular}{l ccc ccc ccc}
\toprule
& \multicolumn{3}{c}{\textbf{Triplet}}
& \multicolumn{3}{c}{\textbf{Relation}}
& \multicolumn{3}{c}{\textbf{Object}} \\
\cmidrule(lr){2-4} \cmidrule(lr){5-7} \cmidrule(lr){8-10}
\textbf{Model}
& VidOR (48s) & Epic-Kitchens (120s) & Ego4D (166s)
& VidOR (48s) & Epic-Kitchens (120s) & Ego4D (166s)
& VidOR (48s) & Epic-Kitchens (120s) & Ego4D (166s) \\
\midrule
\ourmodel
& 19.2 & 6.5 & 5.1
& 20.1 & 7.4 & 7.7
& 81.7 & 63.2 & 66.0 \\
\bottomrule
\end{tabular}
}
\end{table}








\newpage
\noindent
\begin{pastelbox}[pastelblue]{Prompt for Structured Parsing}
\captionof{table}{Prompts for LLM Structured Parsing}
\label{tab:prompt_parse}

\footnotesize
\hangindent=2em
\hangafter=1

\textbf{System Prompt}:\\
\#\# You are an expert in scene understanding. I will give you a short paragraph that describes a video clip.\\
\#\#\# Your task is to extract structured information about a single object described in the paragraph.\\
\#\#\# Please return a JSON with the following fields:\\

-"object": The main object being described (e.g.,"person","dog", "car"). If the inference about the object is uncertain based on the description, add "(uncertain)" after the object name.\\

- "attributes": A list of ONLY the visual/physical attributes that can be directly observed about the object itself. Include only:\\
    * Visual appearance: color, shape, size, texture, pattern, material appearance, style\\
    * Physical properties: state, transparency, reflectiveness, orientation, material\\
    * Design elements: stripes, dots, logos, decorative features.
    DO NOT include: implied states, inferred conditions, functional descriptions, or anything that describes the object's interaction with its environment.\\

- "relationships": A list of relationships between this object and other entities or the environment (e.g., "on top of table", "next to person", "inside container", "facing camera", "part of group").\\

- "actions": A list of actions that the object is performing or movements it is making (e.g., "rotating", "moving", "falling", "bouncing", "sliding").\\

\#\#\# Important distinctions:\\
- Attributes = What the object looks like (visual only). Please use ADJECTIVE form \\
- Relationships = How the object relates to other things spatially, functionally, or contextually \\
- Actions = What the object is doing or how it's moving \\

Now process the following description:

[description]
\end{pastelbox}

\newpage
\noindent
\begin{pastelbox}[pastelblue]{Prompt for Spatial Relationship Extraction}
\captionof{table}{Prompt for Spatial Relationship Extraction}
\label{tab:prompt_spatial}
\footnotesize
\hangindent=2em
\hangafter=1
\textbf{System Prompt}: 

\#\# Role\\
You are a detail-oriented **Video Relationship Annotator** responsible for reviewing sequences of sampled video frames and extracting a comprehensive set of **spatial relationships**. All relationships must be visually grounded, type-consistent, and strictly follow the defined schema.

Begin with a concise checklist (3–7 bullets) of what you will do; keep items conceptual, not implementation-level.

\#\# Task Context\\
- Videos are sampled at 1 fps.\\
- Each frame contains detected objects with bounding boxes and corresponding unique integer IDs.\\
- Your analysis should consider all sampled frames jointly and produce relationships in accordance with the system instructions.\\

\#\# Input Format\\
The input is an ordered sequence of (image URL, text) pairs:\\
- Each text entry includes:\\
  - Frame ID (integer index),\\
  - Frame size (width and height),\\
  - Detected objects (unique integer IDs, labels, and bounding boxes '[x1, y1, x2, y2]').\\
- Frames maintain consistent object IDs when possible. If an object is missing in a frame, its ID is absent for that frame.\\

\#\# Guidelines\\
- Objects are always referenced using their bounding boxes '[x1, y1, x2, y2]' in each frame; base all reasoning on these and their temporal evolution.\\
- Extract **only purely spatial (physical or geometric) relationships** visible in the 3D scene. Do **not** extract any relationships that indicate state, function, action, or purpose.\\
- **Exclude** all temporal, social, functional, or attentional relationships, as well as any stateful or action-based verbs.\\
- Each relationship must be visually supported by the frames.\\
- Do **NOT** ignore objects that newly appear in the middle of the video; ensure to evaluate all newly-detected objects for relationships as soon as they appear.\\
- Ensure logical consistency with common sense and real-world physics; do **NOT** output implausible or unsupported relationships.\\
- Think in terms of **3D spatial layout** by using depth information derived from world knowledge and visual cues, not just 2D image positions. Do **NOT** rely solely on 2D bounding box coordinates. Do **NOT** output 'left of' or 'right of'.\\
- Use precise, explicit, and non-redundant verbs.\\
- Do **NOT** miss any clear and valid spatial relationships between objects.\\

\#\#\# Temporal Grounding\\
- Output relationships with one or more time spans ([[start\_frame, end\_frame], ...]), as continuous intervals where the relationship is visually supported and both objects are present.\\

\#\# Output Format\\
Return a single valid JSON object following this schema:\\
\begin{verbatim}
{
  "relationships": [
    [subject_id, 
    predicate_verb, 
    object_id, 
    [[start_frame, end_frame], ...]]
  ]
}
\end{verbatim}\\

\#\#\# Output Specifications\\
- Each relationship is a tuple with 4 elements, in order: subject\_id (int) | predicate\_verb (str) | object\_id (int) | time\_frames (list of [start,end]).\\
- If there are no valid relationships, output { "relationships": [] }.\\
- Do not emit error fields or any content outside the defined schema in your output.\\
- Always output strictly formatted, valid JSON with the "relationships" key.\\

After producing the output, validate that all relationships are visually supported, ids and frames are valid integers, and the JSON format strictly matches the schema. If validation fails, self-correct before finalizing the output.

\end{pastelbox}

\newpage
\noindent
\begin{pastelbox}[pastelblue]{Prompt for Non-spatial Relationship Extraction}
\captionof{table}{Prompts for Non-spatial Relationship Extraction.} 
\label{tab:prompt_temporal}
\footnotesize
\hangindent=2em
\hangafter=1

\textbf{System Prompt}: \\
\#\# Role:
You are a detail-oriented **Video Relationship Annotator** tasked with reviewing sequences of sampled video frames and extracting a comprehensive set of **temporal (non-spatial) relationships**. Ensure that all extracted relationships are visually grounded, type-consistent, and strictly adhere to the defined schema.
You are analyzing videos sampled at 1 fps. Each frame contains detected objects with corresponding bounding boxes. Please analyze all frames jointly and output temporal relationships according to the system instructions.

\#\# Relationship Taxonomy:
Classify each relationship into exactly ONE category:

1. **Functional — Contact / Manipulation**\\
   - Direct physical interaction where an animate subject alters or uses the state of another object.\\
   - Subject: animate | Object: animate or inanimate\\
   - Exclude: Pure motion or gaze without contact.

2. **Stateful — Attachment / Possession-like**\\
   - Visually grounded, time-persistent attachment or carrying relationships that indicate sustained physical association rather than instantaneous action. \\
   - Subject: animate or inanimate | Object: animate or inanimate\\
   - Exclude: Abstract ownership or purely spatial layout.

3. **Motion — Relative Movement**\\
   - Temporal changes in relative position or movement trajectory between entities.\\
   - Subject: movable (animate or movable inanimate) | Object: animate or inanimate\\
   - Exclude: Static layout or manipulation actions.

4. **Social — Animate-to-Animate Interaction**\\
   - Communication, coordination, or interpersonal acts between animate agents.\\
   - Subject/Object: animate\\
   - Exclude: One-sided attention or non-social contact.

5. **Attentional — Gaze / Focus (includes Camera)**\\
   - Visual attention or camera focus directed at another object or agent.\\
   - Subject: animate or camera (with object\_id = -1) | Object: animate or inanimate\\
   - Exclude: Communication or manipulation.\\
   - **For the camera (object\_id = -1), only extract main or relevant relationships such as its movement or meaningful interaction with other objects; ignore obvious relationships such as "observing" or "recording" unless they are non-trivial or central to the event.**

6. **Event-Level — Goal-Directed Multi-Step Activity**\\
   - Higher-level, time-extended actions combining multiple functional, causality or motion relations into a single purposeful event. \\
   - Subject: animate or inanimate | Object: animate or inanimate\\
   - Exclude: Single short actions or ungrounded intent.

\#\# Core Analysis Logic \& Constraints\\
1. **Object Typing:**
   - Animate: humans, animals, humanoid robots
   - Inanimate: cars, tools, furniture, etc.
   - Camera: unseen observer/recorder, always object\_id = -1
   
2. **Typing Rules:**
   - Functional / Social: Subject must be animate
   - Motion: Subject must be movable
   - Attentional: Subject must be animate or camera (-1)
   - Social: Both subject and object must be animate
   
3. **Extraction Basis:**

   - All relationships must be visually supported and logically consistent with common sense. Do **NOT** infer relationships not visually evidenced or that contradict common sense.
   
   - Every object must be referenced by its bounding box coordinates [x1, y1, x2, y2] in each frame; all reasoning must rely on these spatial positions, object names and their temporal changes.
   
   - Relationships must have temporal grounding: define one or more time spans [[start\_frame, end\_frame], ...] as continuous frame intervals supported by visual evidence. If an object in a labeled relationship disappears from subsequent frames, end the relationship at the object's last visible frame. Do not continue relationships if objects become occluded or are missing.
   
   - Do NOT create self-relations (subject\_id == object\_id), e.g. (i, verb, i, ...).
   
   - If two object IDs correspond to the same physical entity, do not report relationships between them.
   
   - Do not ignore objects that newly appear in the middle of the video; evaluate all newly-detected objects for relationships as soon as they appear.

\#\# Input Format

The input is an ordered sequence of (image URL, text) pairs:
- Each text entry gives the frame ID, frame size (width and height), and detected objects with unique IDs, labels, and bounding boxes [x1, y1, x2, y2].
- Frames are sampled at 1 fps and share consistent object IDs.

\#\# Output Format:
Produce a single valid JSON object using this schema:
\begin{verbatim}
{
  "relationships": [
        [subject_id, 
        predicate_verb, 
        object_id, 
        [[start_frame, end_frame], ...], 
    relationship_type]
  ]
}
\end{verbatim}
\#\#\# Output Details

- Each tuple must have five elements in this order:
subject\_id, predicate\_verb, object\_id, time\_frames (list of (start,end)), relationship\_type (str $\in$ functional, stateful, motion, social, attentional, event\_level)
- If there are no valid relationships, output { "relationships": [] }.
\end{pastelbox}

\end{document}